\documentclass{article}

\usepackage[main, final]{neurips_2026}

\usepackage[utf8]{inputenc} 
\usepackage[T1]{fontenc}    
\usepackage{hyperref}       
\usepackage{url}            
\usepackage{booktabs}       
\usepackage{amsfonts}       
\usepackage{nicefrac}       
\usepackage{microtype}      
\usepackage{xcolor}         

\usepackage{amsmath}
\usepackage{amssymb}
\usepackage{mathtools}
\usepackage{amsthm}
\usepackage{wrapfig}
\usepackage{multirow}
\usepackage{array}
\usepackage{wrapfig}
\usepackage{subcaption}
\usepackage{enumitem}
\usepackage{subcaption}

\newcommand{\name}{\textsc{ReGDiff}}

\usepackage[capitalize,noabbrev]{cleveref}

\theoremstyle{plain}

\theoremstyle{definition}

\theoremstyle{remark}

\usepackage[textsize=tiny]{todonotes}
\newcommand{\eg}{{e.g.}}

\title{\name: Guided Diffusion in Regulated Latent Space for Exploring Metamaterial Voxel Geometry}

\author{%
  Wangzhi Zhan\\
  Department of Computer Science\\
  Virginia Tech\\
  Blacksburg, VA 24061 \\
  \texttt{wzhan24@vt.edu} \\
  \And
  Jianpeng Chen\\
  Department of Computer Science\\
  Virginia Tech\\
  Blacksburg, VA 24061 \\
  \texttt{jianpengc@vt.edu} \\
  \AND
  Dongqi Fu\\
  Meta\\
  Menlo Park, CA 94025 \\
  \texttt{dongqifu@meta.com} \\
  \And
  Dawei Zhou\\
  Department of Computer Science\\
  Virginia Tech\\
  Blacksburg, VA 24061 \\
  \texttt{zhoud@vt.edu} \\
}

\begin{document}

\maketitle

\begin{abstract}
Metamaterials are artificially engineered structures whose mechanical and physical behaviors are strongly shaped by geometry rather than composition. Voxel representation provides a unified format for metamaterial geometry generation, as it can express diverse classes such as truss, shell, and porous structures within a single cubic discretization. However, voxel-based generation faces a plausibility--novelty trade-off: staying close to known geometries helps preserve geometric regularities, while moving away from them is necessary for novelty but may produce degenerate geometries. To address this challenge, we propose \textbf{\name}, a generative framework that couples voxel representation with \emph{latent space regulation} and \emph{guided diffusion}. \name~introduces a repel-and-sink (RAS) mechanism to smooth the latent distribution of plausible geometries, and short-range repulsion (SRR) guidance to discourage generation overly close to known samples while maintaining geometric plausibility. We further contribute a voxel-based benchmark covering truss- and shell-type metamaterial geometries, together with an evaluation module for geometric plausibility, novelty, and diversity. Experiments show that \name~outperforms voxel-based generative baselines, achieving +8.9\% in geometric plausibility, +46.4\% in novelty, and +128.6\% in diversity on average across two datasets. These results suggest that \name~is a strong geometry candidate generator for downstream evaluation. Our code is provided at \url{https://github.com/wzhan24/ReGDiff}.
\end{abstract}

\section{Introduction}
Metamaterials are artificially engineered structures whose unusual behaviors arise from carefully designed geometries rather than intrinsic chemical composition. This structural programmability enables properties rarely observed in natural materials, such as negative Poisson's ratio, ultrahigh stiffness-to-weight ratio, and extreme energy absorption~\citep{nega_poisson, mizzi2020lightweight}. These capabilities have driven breakthroughs across domains including biomedical scaffolds, vibration isolation, acoustic cloaking, soft robotics, and thermal management~\citep{bertoldi2017flexible, liu2011metamaterials}. The ability to tailor functionality through geometry positions metamaterials as a critical frontier for next-generation engineering systems.

Given their extraordinary potential, metamaterials have become a rising focus in material science over the past two decades~\citep{kadic20193d}. Early geometry construction efforts relied heavily on human expertise and manual design, but the emergence of machine learning has enabled data-driven approaches for generating and screening candidate structures. Existing methods largely fall into two categories: modeling metamaterials as 3D graphs~\citep{UniMate, Stiffness, maurizi2025designing}, or designing 2D patterns that are extruded uniformly along a third axis~\citep{kollmann2020deep, tian2022machine, wilt2020accelerating}. Graph representations provide an abstract and interpretable view of metamaterials, yet they lack the ability to express fine-grained geometric details, as edges are usually instantiated as simple cylinders. In contrast, 2D pattern-based designs construct a repeating planar motif and then extend it uniformly along the third axis to form a 3D structure. Such designs can achieve superior performance in the two in-plane directions defined by the patterned motif, but along the extended axis the properties remain largely unchanged from the base material.

Recently, voxel representation, i.e., discretizing a cubic space into small cells marked as void or solid, has become an emerging direction for metamaterial geometry generation. Unlike representations tailored to specific classes of metamaterials, such as graphs for trusses or images for 2D patterns, voxel representation provides a unified and fine-grained format that can express diverse metamaterial geometries, including truss-based, shell-based, porous, 2D, and kirigami structures. This makes voxel representation a compelling modality for geometry candidate generation and evaluation. However, only a few attempts~\citep{zheng2023deep,3D-CDM,yang2024guided} have explored this direction, often by directly adapting 3D generative models from the computer vision domain to metamaterial geometries without explicitly considering the structural regularities expected in metamaterial unit cells.

Despite its promise, voxel-based metamaterial geometry generation faces a key plausibility--novelty trade-off. Generated candidates must satisfy basic geometric plausibility requirements, including symmetry, periodic boundary consistency, and connected solid regions. At the same time, they should not remain too close to training geometries. Existing voxel-based approaches remain limited in this respect: diffusion-based methods~\citep{3D-CDM, yang2024guided} and generative adversarial models~\citep{zheng2023deep} often approximate the observed geometry distribution directly, which can bias generation toward memorized samples; when pushed toward more aggressive exploration, they may instead produce degenerate structures such as disconnected fragments, or even pure voids. This motivates a generation framework that can encourage novelty while maintaining geometric plausibility.


Formally, we identify two key challenges for voxel-based metamaterial geometry generation. \textbf{C1. Plausibility--Novelty Trade-off:} generated candidates should be sufficiently different from known samples while still satisfying basic geometric constraints, including symmetry, periodicity, and connectivity. \textbf{C2. Lack of Benchmark:} to the best of our knowledge, only Yang et al. 2024~\cite{yang2024guided} provides a public large-scale voxel dataset of shell-type metamaterials suitable for training deep generative models. However, many important metamaterial families, such as truss-based structures, are not yet covered in voxel form. Moreover, existing evaluations are often based on visualization or limited structural checks, and a systematic benchmark for voxel-based metamaterial geometry generation is still lacking.

To address \textbf{C1}, we propose \textbf{\name}, a generative framework that combines latent \textbf{Re}gulation with \textbf{G}uided \textbf{Diff}usion. \name~encodes voxel structures into a low-dimensional latent space via an autoencoder, then applies a repel-and-sink (RAS) mechanism to separate plausible geometries from perturbed degenerate ones while smoothing the latent distribution of plausible samples. To further encourage novelty, we introduce short-range repulsion (SRR) guidance into the diffusion process, which discourages generation overly close to known samples while keeping the guidance local to avoid pushing samples outside plausible regions. To address \textbf{C2}, we construct, to the best of our knowledge, the first publicly available large-scale voxel dataset for truss-based metamaterial geometries and propose five metrics to jointly evaluate geometric plausibility, novelty, and diversity.

Through extensive experiments on our dataset and the dataset from~\cite{yang2024guided}, we show that \name~outperforms voxel-based generative baselines, improving geometric plausibility by $8.9\%$, novelty by $46.4\%$, and diversity by $128.6\%$ on average across both datasets. Additional analyses and visualizations of the latent space and generated structures further verify the effectiveness of RAS for latent regulation and SRR for guided generation.

In this work, we make the following contributions:

\vspace{-1em}
\begin{itemize}
  \item \textbf{Framework.} We propose \name, combining RAS latent regulation
  with SRR-guided latent diffusion for structurally plausible and novel voxel
  metamaterial geometry generation.

  \item \textbf{Benchmark.} We build a systematic voxel benchmark covering data
  and evaluation: we release \textbf{MetaTruss} together with \textbf{MetaShell}, and provide metrics for geometric plausibility, novelty,
  and diversity.

  \item \textbf{Experiments.} We conduct comprehensive geometry-level experiments
  showing that \name~better balances geometric plausibility, novelty, and
  diversity than voxel-based generative baselines.
\end{itemize}

\vspace{-1 em}
\section{Preliminaries}
\vspace{-0.3em}

\begin{figure}[t]
    \centering
    \begin{subfigure}[t]{0.52\columnwidth}
        \centering
        \includegraphics[width=\linewidth]{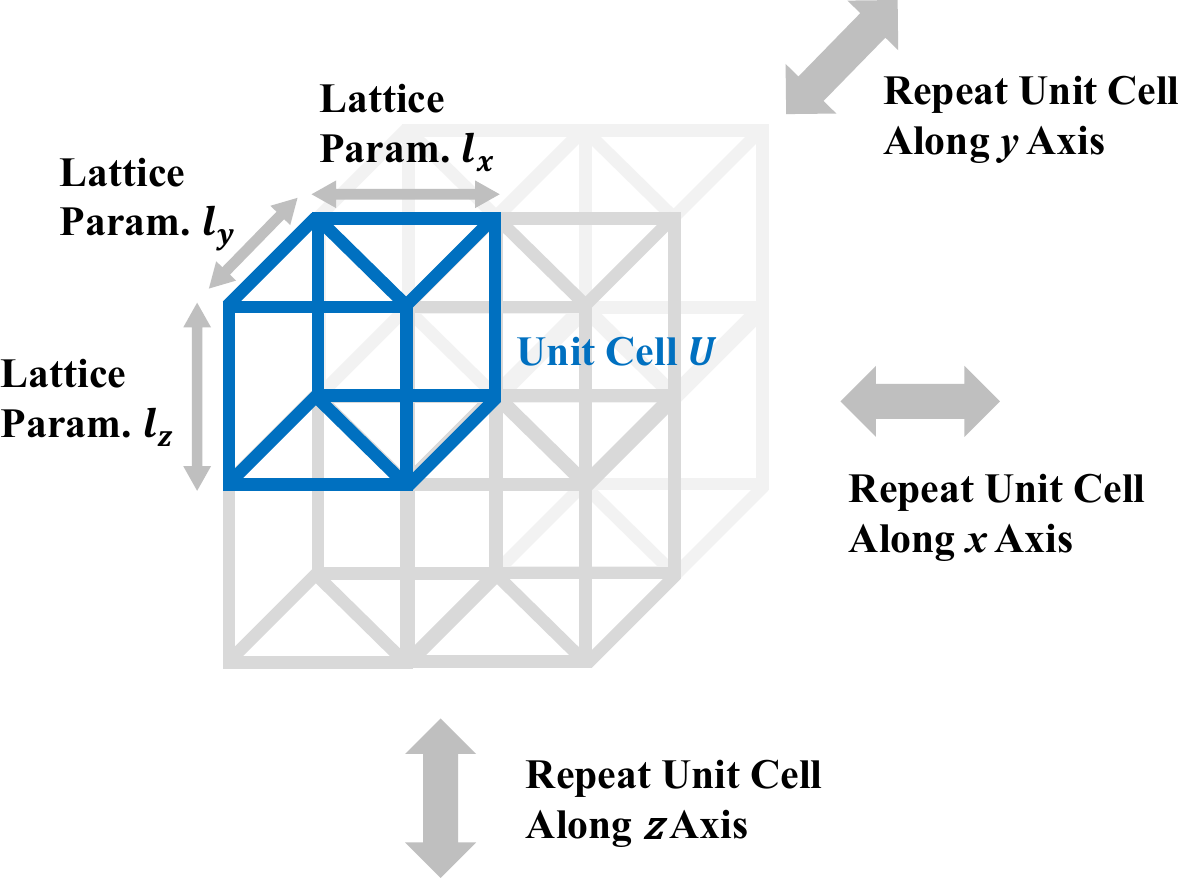}
        \caption{Unit cell and lattice of metamaterials.}
        \label{fig:metamat_illu}
    \end{subfigure}
    \hfill
    \begin{subfigure}[t]{0.4\columnwidth}
        \centering
        \includegraphics[width=\linewidth]{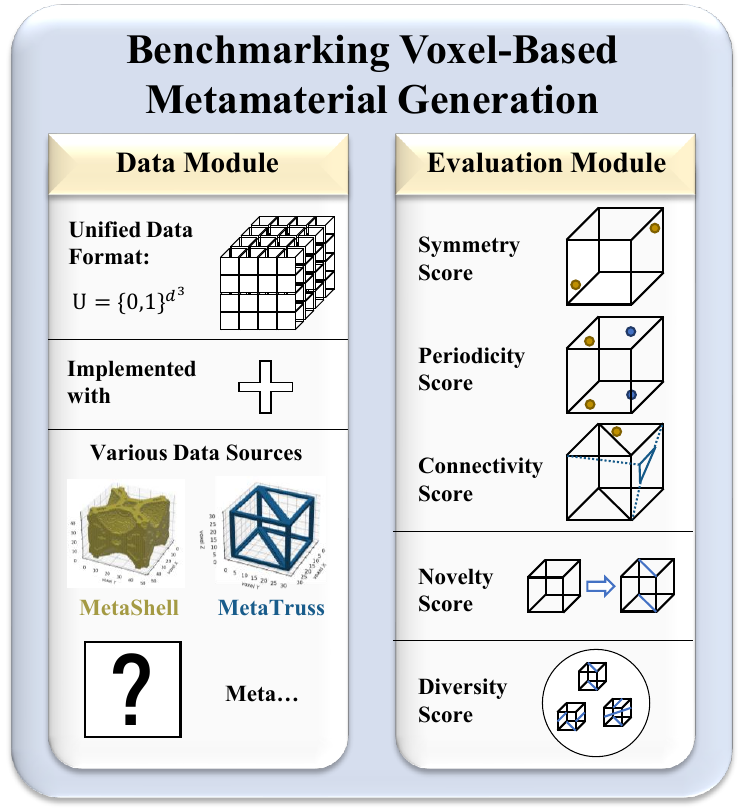}
        \caption{Benchmark development.}
        \label{fig:bench}
    \end{subfigure}

    \vspace{-0.5em}
    \caption{Overview of metamaterials and benchmark development.}
    \label{fig:metamat_bench}
    \vspace{-1em}
\end{figure}

This section introduces metamaterial geometry representation, related voxel generation models, and the geometry candidate generation problem studied in this work.

\subsection{Voxel-Representation for Metamaterials}
\vspace{-0.3em}
Metamaterials are artificial micro-structures composed of substrate material (\eg, plastics, metals, ceramics). Their geometry can be naturally described by a unit cell $\mathbf{U}$ and a lattice vector $\boldsymbol{l}=(l_x,l_y,l_z)\in\mathbb{R}^3$, where $\mathbf{U}$ defines the spatial distribution of substrate material within a cube or cuboid unit, and $\boldsymbol{l}$ specifies repetition intervals along the $x$, $y$, and $z$ axes (Figure~\ref{fig:metamat_illu}). A metamaterial geometry can therefore be denoted as $\mathcal{M}=(\mathbf{U},\boldsymbol{l})$.  
In this work, we focus on generating unit-cell geometries. Specifically, $\mathbf{U}$ is expressed in voxel form as a binary tensor $\mathbf{U}\in\mathbb{B}^{d^3}$, where $\mathbb{B}=\{0,1\}$ and $d$ is the voxel resolution. We denote the dataset of voxelized unit-cell samples as $\mathcal{U}$.

\subsection{Related Models for Voxel Generation}
\vspace{-0.3em}
\paragraph{Autoencoders (AEs).} An AE maps voxel data to a latent space via an encoder $\mathcal{E}$ and reconstructs it with a decoder $\mathcal{D}$. Training minimizes reconstruction loss:

\vspace{-1em}
{\small
\begin{equation}
\label{eq:recon_loss}
L_{\mathrm{recon}}=\frac{1}{N}\sum_{i=1}^{N}||\mathcal{D}\circ\mathcal{E}(\mathbf{U}_i)-\mathbf{U}_i||,
\end{equation}
\vspace{-1em}}

where $\mathbf{U}_i$ is the $i$th voxel sample, $N$ is the dataset size, and $\cdot\circ\cdot$ denotes function composition. The latent variable is $\boldsymbol{x}_i=\mathcal{E}(\mathbf{U}_i)$. To enable generation, the latent distribution of $\boldsymbol{x}$ must be specified or approximated. For instance, variational AEs (VAEs~\cite{vae}) regularize $\boldsymbol{x}$ to follow a Gaussian distribution and sample $\boldsymbol{x}\sim \mathcal{N}(0,1)$ for decoding.

\paragraph{Diffusion Models (DMs).} DMs connect arbitrary data distributions with Gaussian noise through reverse denoising. Following DDPM~\citep{ddpm}, each denoising step can be expressed as:

\vspace{-1em}
{\small
\begin{equation}
\label{eq:diff}
\boldsymbol{x}_{t-1}=\frac{1}{\sqrt{1-\beta_t}}\left(\boldsymbol{x}_t-\frac{\beta_t}{\sqrt{1-\alpha_t^2}}\phi_\mathrm{diff}(\boldsymbol{x}_t,t)\right)+\rho_t\boldsymbol{\epsilon},
\end{equation}}
\vspace{-1em}

where $\phi_\mathrm{diff}$ is the diffusion model, $\boldsymbol{\epsilon}$ is Gaussian noise, and $\alpha_t$, $\beta_t$, and $\rho_t$ are hyperparameters. DMs can also be applied in latent spaces, commonly referred to as latent DMs~\citep{lat_diff}.

\subsection{Problem Definition}
\vspace{-0.3em}
This work studies geometry candidate generation for voxel-based metamaterials. Given a set of voxelized unit-cell samples, our goal is to learn a generative model that produces candidate geometries balancing three aspects: geometric plausibility, novelty, and diversity. Geometric plausibility measures whether a generated unit cell satisfies basic geometric regularities expected in metamaterial candidates, including symmetry, periodic boundary consistency, and connectivity. Novelty measures how different a generated candidate is from known samples, and is therefore defined with respect to a reference dataset. Diversity measures whether the generated candidates cover varied regions of the geometry space rather than collapsing to a small number of similar structures.
\vspace{-0.3em}
\paragraph{Problem Definition.}
Let $f$ denote a generative model that maps a latent variable $\boldsymbol{x}$ to a voxelized unit cell, i.e., $\mathbf{U}=f(\boldsymbol{x})$. The objective is to identify an $f$ that generates voxel metamaterial candidates with high geometric plausibility, novelty, and diversity.

\vspace{-0.2em}
\section{Benchmark Development}
\vspace{-0.3em}
To the best of our knowledge, \textbf{MetaShell}~\citep{yang2024guided} is the only publicly available large-scale voxel-based metamaterial geometry dataset suitable for training deep generative models. Meanwhile, existing evaluation of generated voxel metamaterial geometries is often based on visualization or human assessment, and a systematic evaluation framework is still lacking. To enable a more comprehensive study of voxel-based metamaterial geometry generation, we propose a benchmark that provides both data support and quantitative evaluation for geometry candidate generation.

\subsection{Dataset Development}

We propose a unified voxel-based representation for metamaterial geometry datasets. \textbf{MetaShell}~\citep{yang2024guided} contains voxel data of shell-type (whose unit cells comprise curved surfaces) metamaterial geometries. While valuable, this dataset covers only one class of geometries. Truss-based metamaterials represent another critical category for mechanical applications~\citep{mizzi2020lightweight, song2025compressive}, yet existing truss datasets rely on graph representations~\citep{MetaModulus, Stiffness}, which lack fine-grained geometric detail. Reformatting truss structures into voxel space not only unifies them with shell-type geometries under a common representation, but also preserves richer geometric detail. To close this gap, we construct a truss-based voxel dataset, which we call \textbf{MetaTruss}. MetaTruss is derived from~\citep{MetaModulus}, where original samples are provided in 3D graph format. Each unit cell is discretized into a $48^3$ voxel grid: voxels lying within a truss radius of any graph edge are marked as solid, while all others remain void. Following this procedure, we process the first 10,000 samples from~\citep{MetaModulus}. With MetaShell also included, our benchmark establishes a unified data module that remains compatible with future metamaterial geometry datasets. More details are in Appendix~\ref{app:bench_detail}.

\subsection{Evaluation Mechanism}
To systematically evaluate generated voxel geometries, we propose five metrics from three aspects. \textbf{Geometric Plausibility Scores:} we use symmetry score $S_\mathrm{sym}$ to evaluate the central symmetry degree of a geometry; periodicity score $S_\mathrm{per}$ to evaluate how similar each facet of the cube frame is to its parallel counterpart; and connectivity score $S_\mathrm{con}$ to evaluate whether the solid voxels form a connected geometry, measured as the volume fraction of the largest connected component. \textbf{Novelty Score:} we use $S_\mathrm{nov}$ to evaluate the IoU (intersection over union) distance between a generated sample and its nearest neighbor in the training dataset. \textbf{Diversity Score:} we use $S_\mathrm{div}$ to evaluate how many different training samples serve as nearest neighbors of generated samples, normalized by the number of generated samples. More details regarding the benchmark can be found in Appendix~\ref{app:bench_detail}.

\begin{figure*}[ht]
    \centering
    \includegraphics[width=0.9\textwidth]{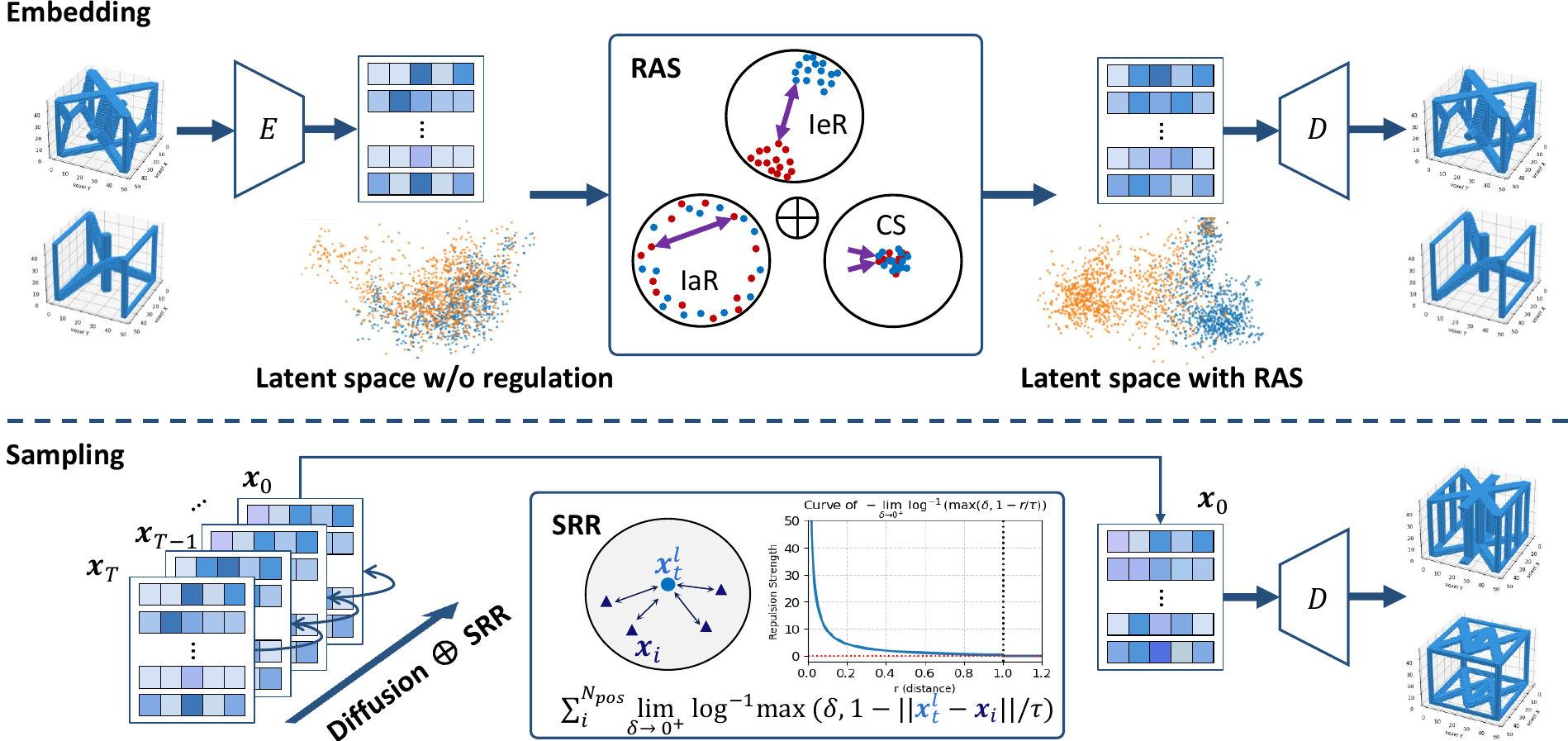}
    \caption{An overview of the proposed framework \name. It encodes voxel geometries into latent space, applies RAS for latent regulation, and employs SRR-guided diffusion to generate novel yet geometrically plausible metamaterial geometry candidates.}
    \vspace{-1.5em}
    \label{fig:framework}
\end{figure*}

\vspace{-0.2em}
\section{Methodology}
\vspace{-0.3em}
This section introduces \name, our framework for voxel-based metamaterial geometry generation. We first give an overview, then present the autoencoder, RAS latent regulation, and SRR-guided diffusion for novel candidate generation.

\subsection{Framework Overview}
The main challenge in voxel-based metamaterial geometry generation is the \textbf{plausibility--novelty trade-off (C1)}: candidates that stay too close to known samples may preserve geometric plausibility but offer limited novelty, while candidates that move too far away may become geometrically degenerate. \name~tackles this challenge in two steps.
First, an autoencoder maps voxels into a compact latent space regulated by the RAS mechanism, which separates plausible geometries from perturbed degenerate ones and smooths the latent distribution of plausible samples. This reduces the sensitivity of decoding to small latent perturbations and improves the robustness of geometry generation.
Second, a latent diffusion model with SRR guidance discourages samples from staying overly close to known geometries, promoting novel candidate generation while keeping the guidance local.
Together, RAS provides a geometry-aware latent space and SRR encourages novelty, enabling \name~to balance geometric plausibility, novelty, and diversity.

\subsection{Autoencoding with RAS Latent Regulation}
To mitigate the high dimensionality of voxel representation, we use an AE to compress voxel data into a low-dimensional latent space.
However, simply regularizing the latent distribution, as in VAEs, often reduces geometric plausibility, since metamaterial unit cells are expected to satisfy basic geometric regularities such as periodic boundary consistency and connectivity. Even slight deviations, such as isolated floating clusters, may lead to geometrically degenerate candidates.
Prior AE-based methods lack tailored latent regulation, resulting in a latent space where plausible and perturbed degenerate geometries can be entangled. Such entanglement may lead to poorly formed generated candidates.
To address this issue, we propose the RAS mechanism, which separates plausible and perturbed degenerate regions in the latent space through three component mechanisms: \textbf{inter-class repulsion (IeR)}, \textbf{intra-class repulsion (IaR)}, and \textbf{central sink (CS)}.
We first synthesize negative voxel samples by perturbing ground-truth geometries for the purpose of this separation.
Let $\mathcal{U}_\mathrm{pos}$, $\mathcal{U}_\mathrm{neg}$, and $\mathcal{U}=\mathcal{U}_\mathrm{pos}\cup\mathcal{U}_\mathrm{neg}$ denote the positive, negative, and full datasets.
Encoding $\mathcal{U}$ with $\mathcal{E}$ yields latent datasets $\mathcal{X}$, with $\mathcal{X}_\mathrm{pos}$ and $\mathcal{X}_\mathrm{neg}$ denoting the positive and negative subsets.

\begin{wrapfigure}{r}{0.5\textwidth}
    \centering
    \vspace{-1.0em}
    \includegraphics[width=0.48\textwidth]{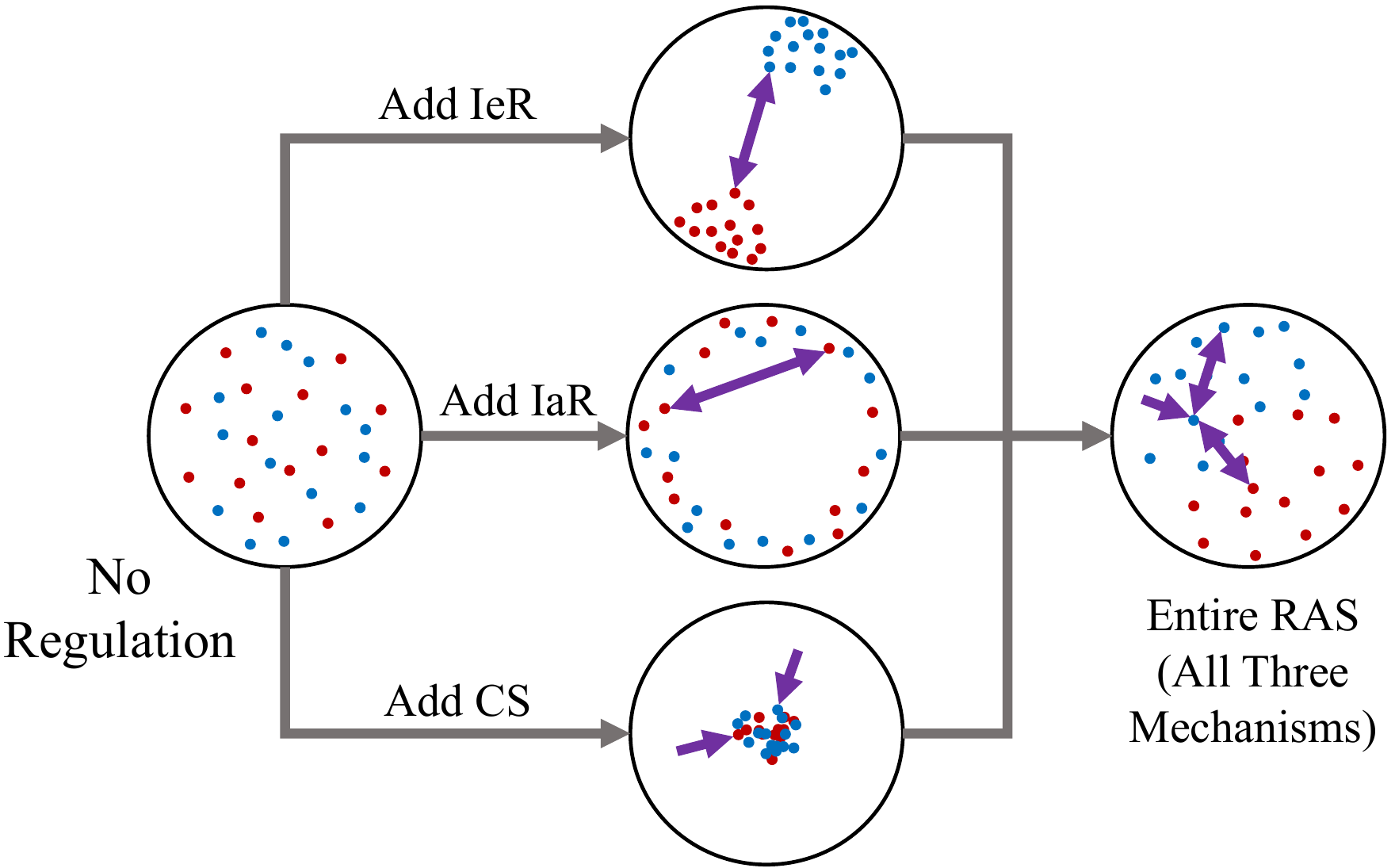}
    \vspace{-0.6em}
    \caption{Illustration of the effect of RAS and its components.}
    \label{fig:ras_mechan}
    \vspace{-1.0em}
\end{wrapfigure}

\paragraph[Inter-Class Repulsion]{Inter-Class Repulsion.\footnote{In this paper $|\cdot|$ means the sum of all elements if a tensor is inside (\eg, $\mathbf{U}$), or the cardinality of a set when a set is inside (\eg, $\mathcal{U}$).}} IeR aims to simplify the decision boundary between $\mathcal{X}_\mathrm{pos}$ and $\mathcal{X}_\mathrm{neg}$ by adding inverse-square repulsion similar to Coulomb repulsion~\citep{coulomb}:

\vspace{-1em}
{\small
\begin{equation}
F_{\mathrm{inter}}(\mathcal{X}_\mathrm{pos},\mathcal{X}_\mathrm{neg})=\sum_{i=1}^{|\mathcal{X_{\mathrm{pos}}}|}\sum_{j=1}^{|\mathcal{X}_{\mathrm{neg}}|}\frac{\boldsymbol{x}_{\mathrm{pos},i}-\boldsymbol{x}_{\mathrm{neg},j}}{||\boldsymbol{x}_{\mathrm{pos},i}-\boldsymbol{x}_{\mathrm{neg},j}||^3},
\end{equation}}
\vspace{-1em}

where $\boldsymbol{x}_{\mathrm{pos},i}$ and $\boldsymbol{x}_{\mathrm{neg},j}$ are the $i$th positive latent sample and $j$th negative latent sample, respectively. The simulated distribution of adding IeR alone can be found in Figure~\ref{fig:ras_mechan}. To optimize the latent distribution with IeR, we minimize the integral of $F_\mathrm{inter}$, i.e., the Coulomb potential:

\vspace{-1em}
{\small
\begin{equation}
P_\mathrm{inter}(\mathcal{X}_\mathrm{pos},\mathcal{X}_\mathrm{neg})=\sum_{i=1}^{|\mathcal{X_{\mathrm{pos}}}|}\sum_{j=1}^{|\mathcal{X}_{\mathrm{neg}}|}||\boldsymbol{x}_{\mathrm{pos},i}-\boldsymbol{x}_{\mathrm{neg},j}||^{-1}.
\end{equation}}
\vspace{-1.5em}

\paragraph{Intra-Class Repulsion.} As illustrated in Figure~\ref{fig:ras_mechan}, using IeR alone can simplify the decision boundary, but it drives the two classes into two distant clusters. In this case, only a small portion of the latent space is covered, so the decoder may not decode latents well outside these concentrated regions. To alleviate this issue, unlike contrastive learning which pulls positive latents closer together~\citep{contrastive}, we propose IaR to reduce the converging tendency within each class and encourage broader latent coverage. Similar to IeR, IaR and its potential are:

\vspace{-1em}
{\small
\begin{equation}
F_{\mathrm{intra}}(\mathcal{X}_{\mathrm{pos}})
=
\sum_{\substack{i,j=1\\ i\neq j}}^{|\mathcal{X}_{\mathrm{pos}}|}
\frac{
\boldsymbol{x}_{\mathrm{pos},i}-\boldsymbol{x}_{\mathrm{pos},j}
}{
\left\|
\boldsymbol{x}_{\mathrm{pos},i}-\boldsymbol{x}_{\mathrm{pos},j}
\right\|^3
},
\qquad
P_{\mathrm{intra}}(\mathcal{X}_{\mathrm{pos}})
=
\sum_{\substack{i,j=1\\ i<j}}^{|\mathcal{X}_{\mathrm{pos}}|}
\left\|
\boldsymbol{x}_{\mathrm{pos},i}-\boldsymbol{x}_{\mathrm{pos},j}
\right\|^{-1}.
\end{equation}}
\vspace{-1em}

By substituting the ``pos'' subscript with ``neg'' we can obtain the IaR equations for negative samples.
\vspace{-0.7em}

\paragraph{Central Sink (CS).}
IeR and IaR simplify the latent decision boundary and avoid intra-class convergence, but they can also force latents to be far from each other, making the latent distribution overly sparse. To alleviate this problem, we propose CS to attract all latents toward the origin. Let $\boldsymbol{x}_i$ be the $i$th latent variable, the force and potential of CS are:
{\small
\begin{equation}
F_\mathrm{sink}(\mathcal{X})=\sum_{i=1}^{|\mathcal{X}|} \boldsymbol{x}_i,~ \mathrm{and~}P_\mathrm{sink}(\mathcal{X})=\sum_{i=1}^{|\mathcal{X}|} ||\boldsymbol{x}_i||^2.
\end{equation}}

The effect of adding CS alone is also shown in Figure~\ref{fig:ras_mechan}. Combining all three mechanisms together, the RAS loss can be defined as:
{\small
\begin{equation}
L_{\mathrm{RAS}}
=
\lambda_\mathrm{inter}P_\mathrm{inter}(\mathcal{X}_\mathrm{pos},\mathcal{X}_\mathrm{neg})
+
\lambda_\mathrm{intra}\big(
P_\mathrm{intra}(\mathcal{X}_\mathrm{pos})
+
P_\mathrm{intra}(\mathcal{X}_\mathrm{neg})
\big)
+
\lambda_\mathrm{sink}P_\mathrm{sink}(\mathcal{X}) .
\end{equation}}
\vspace{-1.2em}

where $\lambda_\mathrm{inter}$, $\lambda_\mathrm{intra}$ and $\lambda_\mathrm{sink}$ are hyperparameters. Then we can obtain the total loss function for training the autoencoder:
{\small
\begin{equation}
L_{\mathrm{auto}}=\lambda_{\mathrm{recon}} L_{\mathrm{recon}} + \lambda_{\mathrm{RAS}} L_{\mathrm{RAS}},
\end{equation}}
where $L_{\mathrm{recon}}$ is defined in Equation~\ref{eq:recon_loss}, and $\lambda_\mathrm{recon}$ and $\lambda_\mathrm{RAS}$ are two hyperparameters.

Combining the three mechanisms, RAS regulates the AE latent space so that plausible and perturbed degenerate geometries are better separated. At the same time, the positive latents are encouraged to occupy a smoother distribution while maintaining distances among samples, avoiding latent collapse where all latents converge to a small region.

\vspace{-0.5em}
\subsection{Latent Diffusion with SRR Guidance}
The RAS mechanism regulates the latent distribution so that the latent decision boundary is simplified, making the decoder more robust to variations around the positive latent sample region. In this latent space, we use diffusion models to approximate the latent distribution and connect it with a Gaussian distribution for sampling. However, a vanilla diffusion paradigm such as DDPM, without further guidance, may still favor regions close to known samples because these regions usually have high generation probability. This can limit novelty in generated candidates.

To address this issue, we introduce a repulsion force between the sample being generated and the latents of known samples, guiding the diffusion process away from overly similar regions of the latent space. Crucially, this repulsion must be short-ranged: if it extends too far, the generated sample could be pushed outside the region where the decoder produces geometrically plausible candidates. Building on this idea, we propose the SRR mechanism, which augments the DDPM model with an additional SRR guidance term. Based on Equation~\ref{eq:diff}, the denoising step of an SRR-guided DDPM can be expressed as:

{\small
\vspace{-1.5em}
\begin{align}
\label{eq:srr_diff}
\boldsymbol{x}_{t-1} &= 
\frac{1}{\sqrt{1-\beta_t}}\left(\boldsymbol{x}_t
-\frac{\beta_t}{\sqrt{1-\alpha_t^2}}
\phi_\mathrm{diff}(\boldsymbol{x}_t,t)\right)
+\rho_t\boldsymbol{\epsilon} \nonumber \\
&\quad+
\lambda_{\mathrm{SRR}}\sum_{i=1}^{N_{\mathrm{pos}}}
\frac{\boldsymbol{x}_{\mathrm{pos},i}-\boldsymbol{x}_{t}}{||\boldsymbol{x}_{\mathrm{pos},i}-\boldsymbol{x}_{t}||} 
\lim_{\delta\to0^+}\log^{-1}\max(\delta,\, 1-
||\boldsymbol{x}_{t}-\boldsymbol{x}_{\mathrm{pos},i}||/\tau),
\end{align}
\vspace{-1.5em}
}

where $\lambda_{\mathrm{SRR}}$ and $\tau$ are two hyperparameters controlling the SRR guidance strength. The SRR guidance, corresponding to the last term in Equation~\ref{eq:srr_diff}, introduces an additional force that pushes the latent variable $\boldsymbol{x}_t$ away from known latents. The logarithmic formulation ensures that this repulsion decays rapidly with distance, aligning with the intuition that only nearby samples should influence the current denoising step. In practice, computing distances to all known latents is prohibitively expensive, so we cluster the known latents before training the diffusion model. During denoising, clusters outside the neighborhood of the current latent variable are ignored by the SRR mechanism, reducing computational cost. Importantly, SRR guidance is applied only at inference, while the training scheme follows the standard DDPM paradigm.
\vspace{1em}\\
To summarize, RAS regulates the AE latent space to separate plausible geometries from perturbed degenerate ones and to form a smoother positive latent distribution for candidate generation. In this regulated latent space, SRR-guided diffusion discourages generated latents from staying overly close to known samples, increasing the chance of producing novel and geometrically plausible candidates.

\vspace{-0.2em}
\section{Experiments}
\vspace{-0.5em}
In this section, we evaluate \name~as a geometry candidate generator for voxel-based metamaterial unit cells.
We compare it with voxel-based generative baselines using geometric plausibility, novelty, and diversity metrics, followed by ablation studies, model capacity analysis, and sample visualizations.

\begin{table*}[t]
\centering
\vspace{-0.5em}
\caption{Performance evaluation of different approaches.}
\vspace{-0.5em}
\label{tab:effectiveness}
\setlength{\tabcolsep}{4pt}
\small
\begin{tabular}{@{}lcccccc@{}}
\toprule
\multirow{2}{*}{Approaches} & \multicolumn{4}{c}{Geometric Plausibility Scores} & Novelty Score & Diversity Score\\ 
\cmidrule(lr){2-5} \cmidrule(lr){6-6} \cmidrule(lr){7-7}
& $S_\mathrm{sym}$ $\uparrow$ & $S_\mathrm{per}$ $\uparrow$ & $S_\mathrm{con}$ $\uparrow$ & Mean $\uparrow$ & $S_\mathrm{nov}$ $\uparrow$ & $S_\mathrm{div}$ $\uparrow$ \\ 
\midrule
\multicolumn{7}{c}{\textbf{MetaTruss} (ours)} \\
\midrule
DiT-3D (\cite{dit3d})            & 0.358 & 0.248 & 0.500 & 0.369 & 0.003 & 0.010 \\
Y. Yang et al. (\cite{yang2024guided}) & \textbf{0.800} & \textbf{0.585} & 0.494 & \underline{0.626} & \underline{0.163} & \underline{0.158} \\
XCube (\cite{xcube})             & 0.506 & \underline{0.525} & 0.522 & 0.518 & 0.000 & 0.004 \\
Trellis (\cite{SLat})            & 0.081 & 0.063 & 0.133 & 0.092 & 0.000 & 0.001 \\
3D-CDM (\cite{3D-CDM})           & 0.470 & 0.270 & \textbf{0.999} & 0.580 & 0.000 & 0.001 \\
\name~(ours)                     & \underline{0.718} & 0.487 & \underline{0.969} & \textbf{0.725} & \textbf{0.296} & \textbf{0.420} \\
\midrule
\multicolumn{7}{c}{\textbf{MetaShell}~\citep{yang2024guided}} \\
\midrule
DiT-3D (\cite{dit3d})            & 0.465 & 0.259 & 0.704 & 0.476 & 0.023 & 0.013 \\
Y. Yang et al. (\cite{yang2024guided}) & \underline{0.922} & \underline{0.791} & \underline{0.991} & \underline{0.901} & 0.342 & \underline{0.409} \\
XCube (\cite{xcube})             & 0.522 & 0.523 & 0.526 & 0.524 & 0.000 & 0.005 \\
Trellis (\cite{SLat})            & 0.795 & 0.576 & \textbf{0.999} & 0.790 & 0.103 & 0.032 \\
3D-CDM (\cite{3D-CDM})           & 0.668 & 0.529 & \textbf{0.999} & 0.732 & \textbf{0.390} & 0.113 \\
\name~(ours)                     & \textbf{0.923} & \textbf{0.856} & 0.978 & \textbf{0.919} & \underline{0.380} & \textbf{0.783} \\
\bottomrule
\end{tabular}
\vspace{-1.5em}
\end{table*}

\vspace{-0.4em}
\subsection{Overall Comparison}
\vspace{-0.5em}
We evaluate \name~against five voxel-based generative baselines: DiT-3D~\citep{dit3d}, Yang et al.~\citep{yang2024guided}, XCube~\citep{xcube}, Trellis~\citep{SLat}, and 3D-CDM~\citep{3D-CDM}. The experiments are conducted on our proposed benchmark, which includes the MetaTruss and MetaShell datasets, and the task is voxel-based metamaterial geometry generation. Performance is assessed using five complementary metrics covering three dimensions: geometric plausibility ($S_\mathrm{sym}$, $S_\mathrm{per}$, $S_\mathrm{con}$), novelty ($S_\mathrm{nov}$), and diversity ($S_\mathrm{div}$). All models are trained and evaluated on a single NVIDIA A100 GPU (except XCube whose large model size requires two A100 GPUs).
To compare the candidate generation capability of \name~with other baseline models, we train each model on each of the two datasets and compute the five metrics for the generated geometries. Specific results are shown in Table~\ref{tab:effectiveness}.

Across both MetaTruss and MetaShell datasets, \name~achieves the best balance of geometric plausibility, novelty, and diversity. On MetaTruss, it delivers competitive geometric plausibility while substantially outperforming baselines in novelty and diversity, reducing the memorization tendency observed in prior methods. On MetaShell, it matches or exceeds baseline-level geometric plausibility and nearly doubles diversity. These results suggest that RAS helps maintain geometric plausibility, while SRR encourages generation away from overly similar training samples. Together, they show that \name~better addresses the plausibility--novelty trade-off than other baselines.

\begin{table*}[h] 
\centering
\vspace{-0.5em}
\caption{Latent distribution visualization and generated samples with RAS regulation, contrastive regulation, or no regulation. Reg. denotes regulation, and Contra. denotes contrastive.}
\vspace{-0.5em}
\label{tab:result_vis}
\setlength{\tabcolsep}{4pt}
\small
\begin{tabular}{>{\centering\arraybackslash}m{2.8cm}|>{\centering\arraybackslash}m{5cm}|>{\centering\arraybackslash}m{5cm}}
\toprule
Latent Regulation & PCA of Latent Distribution & Generated Geometries \\ 
\midrule
RAS Reg. & \includegraphics[width=0.9\linewidth]{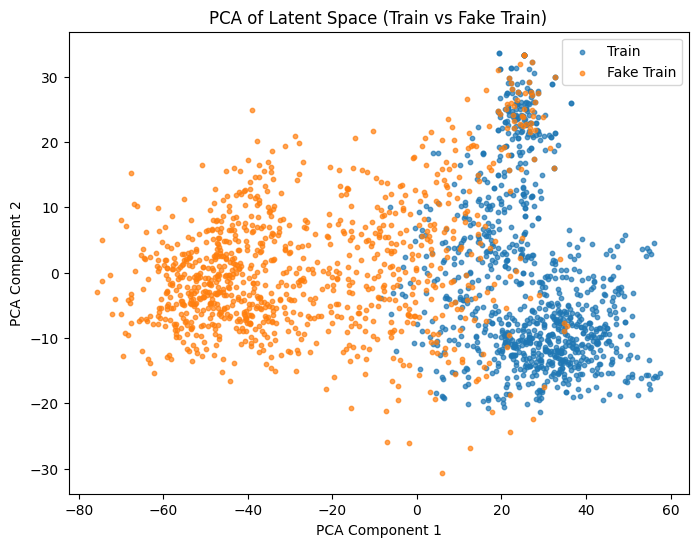} 
         & \includegraphics[width=0.9\linewidth]{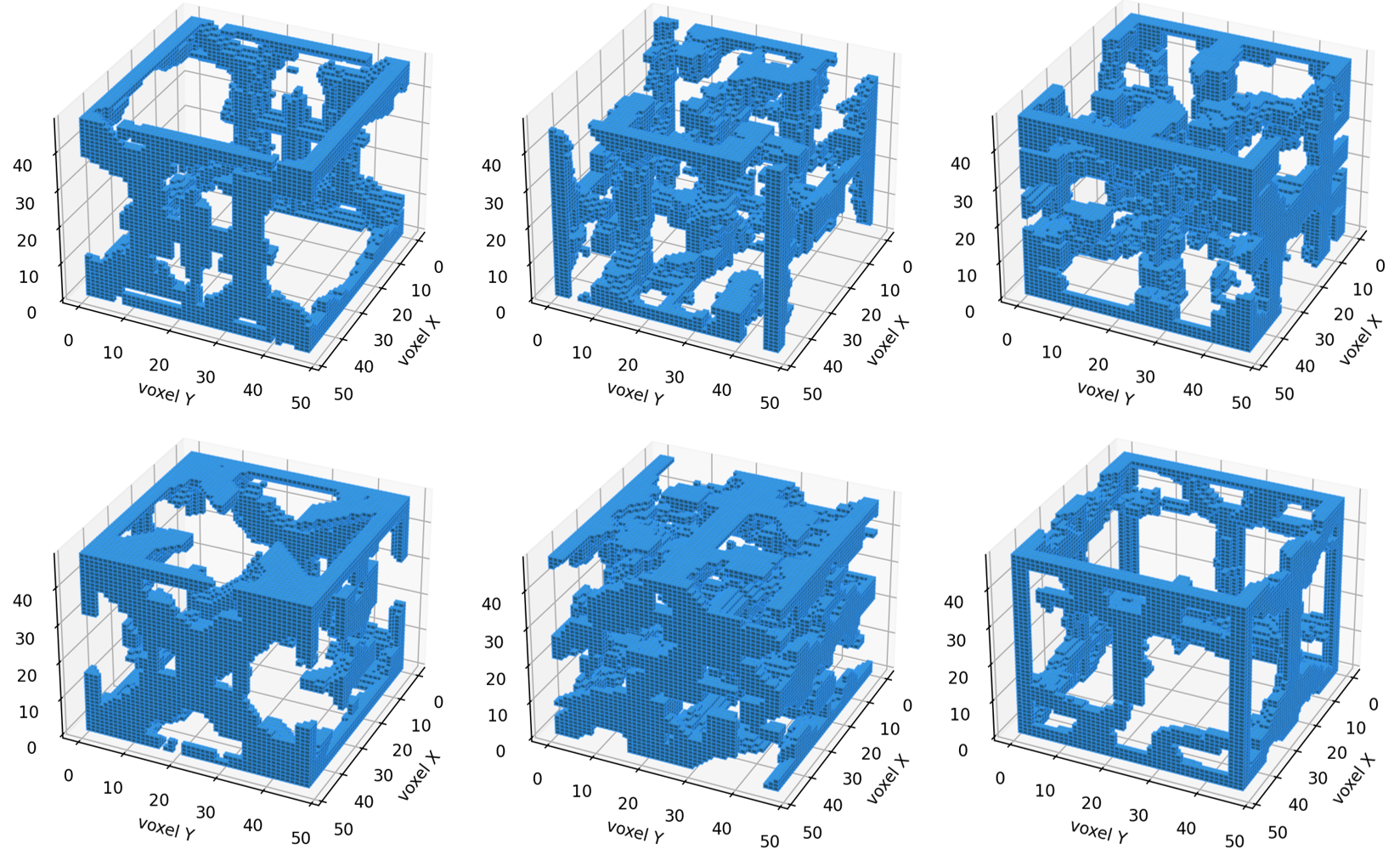} \\
\midrule
Contra. Reg. & \includegraphics[width=0.9\linewidth]{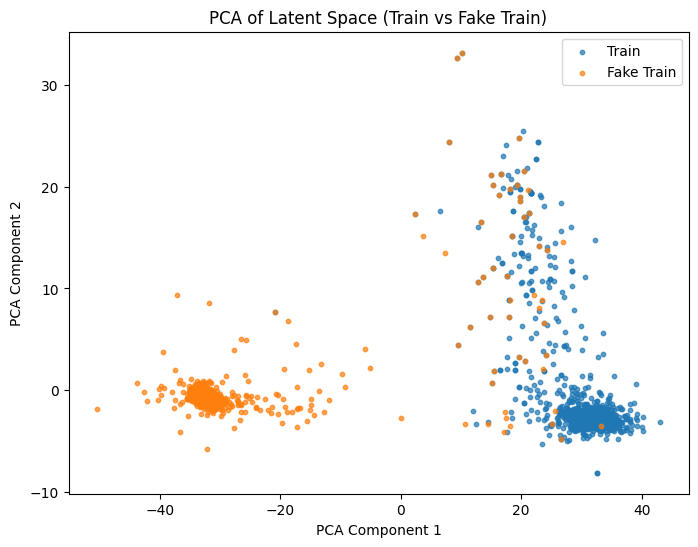} 
            & \includegraphics[width=0.9\linewidth]{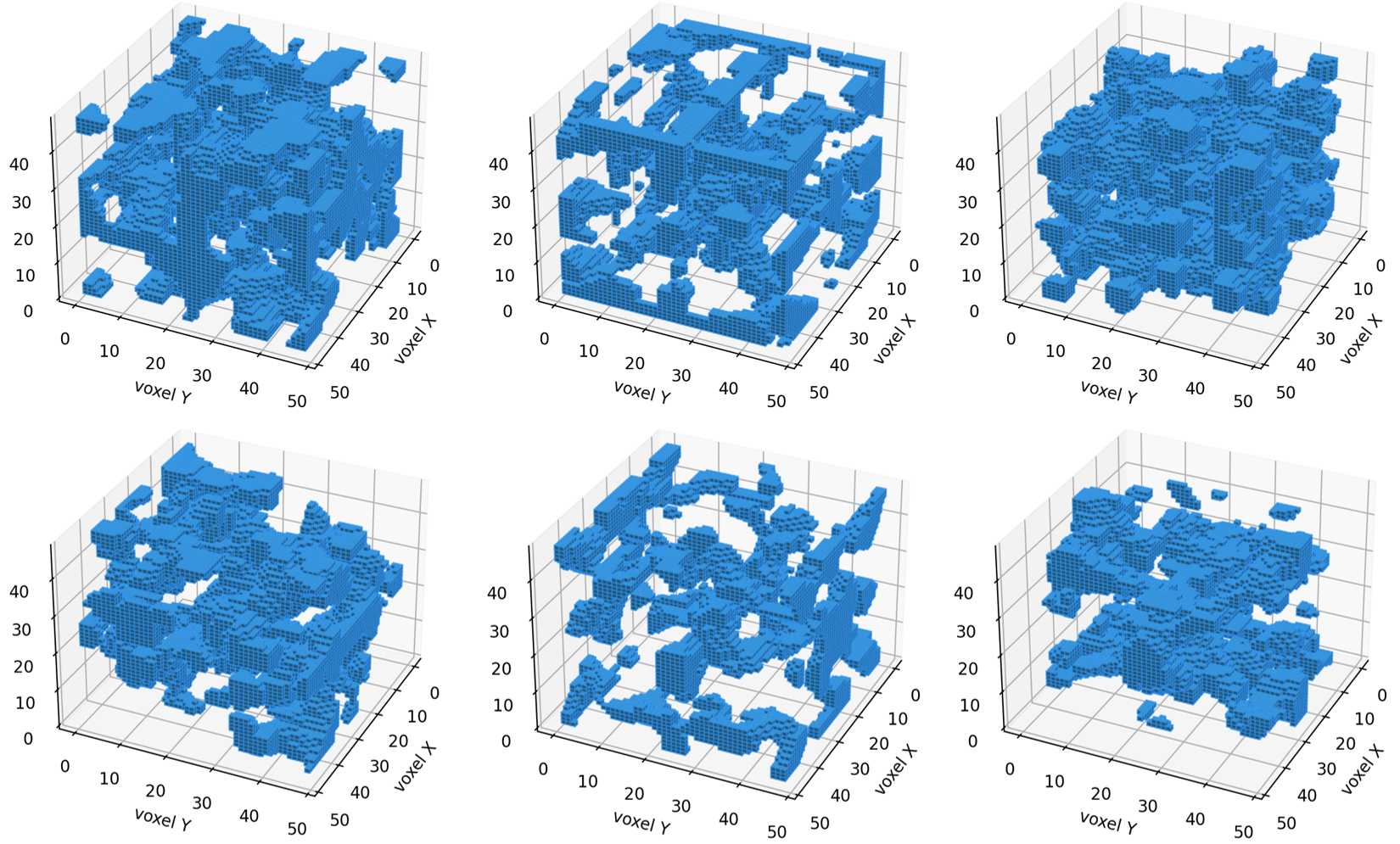} \\
\midrule
w/o Reg. & \includegraphics[width=0.9\linewidth]{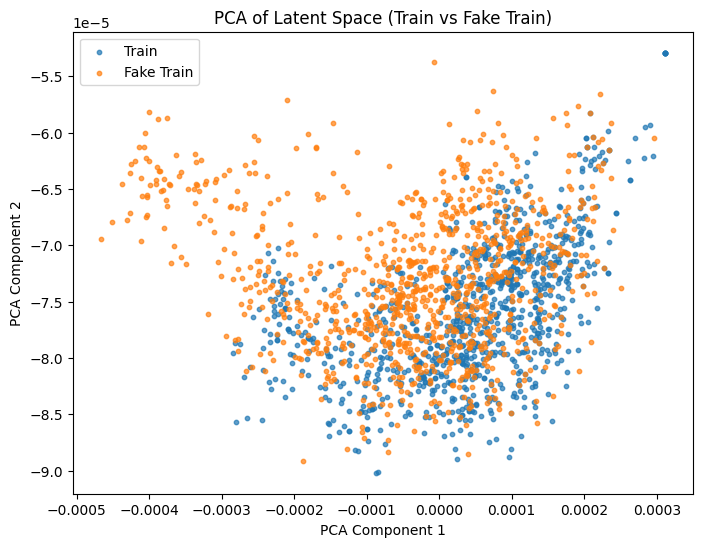} 
         & \includegraphics[width=0.9\linewidth]{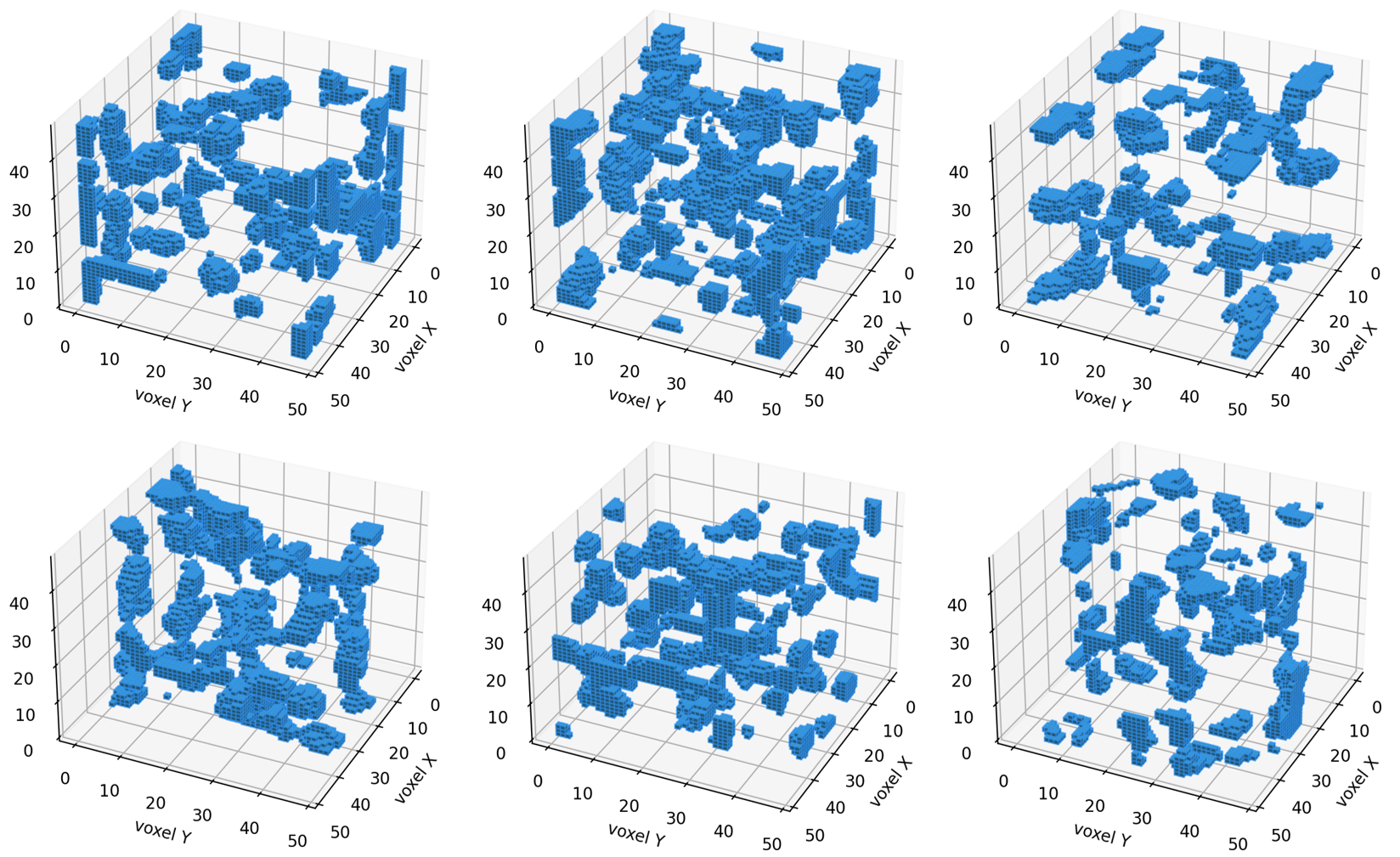} \\
\bottomrule
\end{tabular}
\vspace{-1em}
\end{table*}

\subsection{Ablation Study}
Experiments in this section are conducted on MetaTruss (Results on MetaShell are in Appendix~\ref{app:shell_ablation}).

\vspace{-1em}
\paragraph{Latent Space Regulations Comparison.}
To verify the effect of RAS, we train the autoencoder under three settings: (1) without latent regulation; (2) with contrastive regulation~\citep{contrastive}; and (3) with RAS regulation. We visualize the PCA-compressed latent distributions and generated geometries in Table~\ref{tab:result_vis}. Without regulation, positive and negative latents are mixed, which harms geometric plausibility during generation. Contrastive regulation separates the two classes but pulls each class into a compact region, limiting latent coverage. In contrast, RAS better separates positive and negative latents while maintaining a smoother latent distribution, leading to more diverse and well-formed geometries. The quantitative results in Table~\ref{tab:SRR_effect} further verify its effectiveness.

\begin{table*}[h]
\centering
\vspace{-0.5em}
\caption{Ablation on RAS regulation and SRR diffusion.}
\vspace{-0.5em}
\label{tab:SRR_effect}
\setlength{\tabcolsep}{4pt}
\small
\begin{tabular}{@{}lcccccc@{}}
\toprule
\multirow{2}{*}{Approaches} & \multicolumn{4}{c}{Geometric Plausibility Scores} & Novelty Score & Diversity Score\\ 
\cmidrule(lr){2-5} \cmidrule(lr){6-6} \cmidrule(lr){7-7}
& $S_\mathrm{sym}$ $\uparrow$ & $S_\mathrm{per}$ $\uparrow$ & $S_\mathrm{con}$ $\uparrow$ & Mean $\uparrow$ & $S_\mathrm{nov}$ $\uparrow$ & $S_\mathrm{div}$ $\uparrow$ \\ 
\midrule
Case 1 (RAS + vanilla DDPM)   & \underline{0.753} & \underline{0.632} & \underline{0.885} & \textbf{0.757} & \underline{0.208} & \underline{0.336} \\
Case 2 (w/o reg + SRR Diff.) & \textbf{0.873} & \textbf{0.801} & 0.295 & 0.656 & 0.014 & 0.011 \\
Case 3 (full framework)      & 0.718 & 0.487 & \textbf{0.969} & \underline{0.725} & \textbf{0.296} & \textbf{0.420} \\
\bottomrule
\end{tabular}
\vspace{-0.5em}
\end{table*}

\begin{table*}[h]
\centering
\caption{Ablation on model capacity. Param. Num. denotes parameter number.}
\vspace{-0.5em}
\label{tab:sense}
\setlength{\tabcolsep}{4pt}
\small
\begin{tabular}{@{}lcccccc@{}}
\toprule
\multirow{2}{*}{Approaches} & \multicolumn{4}{c}{Geometric Plausibility Scores} & Novelty Score & Diversity Score\\ 
\cmidrule(lr){2-5} \cmidrule(lr){6-6} \cmidrule(lr){7-7}
& $S_\mathrm{sym}$ $\uparrow$ & $S_\mathrm{per}$ $\uparrow$ & $S_\mathrm{con}$ $\uparrow$ & Mean $\uparrow$ & $S_\mathrm{nov}$ $\uparrow$ & $S_\mathrm{div}$ $\uparrow$ \\ 
\midrule
Increase AE Param. Num.   & \textbf{0.753} & 0.471 & \textbf{0.977} & \textbf{0.734} & \underline{0.308} & 0.411 \\
Decrease AE Param. Num.   & 0.688 & 0.479 & 0.953 & 0.707 & 0.280 & 0.413 \\
Increase diff. Param. Num.& 0.705 & 0.474 & 0.936 & 0.705 & \textbf{0.330} & \textbf{0.444} \\
Decrease diff. Param. Num.& \underline{0.722} & \textbf{0.493} & 0.955 & 0.723 & 0.286 & 0.409 \\
Original setting          & 0.718 & \underline{0.487} & \underline{0.969} & \underline{0.725} & 0.296 & \underline{0.420} \\
\bottomrule
\end{tabular}
\vspace{-1em}
\end{table*}

\vspace{-0.5em}
\paragraph{SRR diffusion vs. vanilla DDPM.}
Our high-level aim is to generate voxel candidates that are both geometrically plausible and novel. To verify the effect of SRR guidance, we compare two cases: (1) RAS + vanilla DDPM model; and (2) RAS + SRR diffusion, corresponding to our full framework. We train the model under these two settings and report the results in Table~\ref{tab:SRR_effect}. From Table~\ref{tab:SRR_effect}, we can see that SRR diffusion provides substantially higher novelty and diversity scores, while keeping the geometric plausibility scores close to vanilla DDPM. The mild decrease in geometric plausibility is expected, because vanilla DDPM tends to generate samples closer to the training distribution, where geometric regularities are easier to preserve. These results show that SRR guidance increases novelty and diversity while maintaining competitive geometric plausibility.

\paragraph{Model Capacity Sensitivity Analysis.}
We further study the effect of model capacity by increasing or decreasing the number of layers in the autoencoder and diffusion backbone (Table~\ref{tab:sense}). Results show that changing the autoencoder depth only slightly alters geometric plausibility, novelty, and diversity, indicating that encoding and decoding are relatively robust to capacity variations. In contrast, modifying the diffusion backbone has a more pronounced impact: adding a layer improves novelty and diversity, while removing a layer reduces both. This suggests that the diffusion model's capacity is more critical than that of the autoencoder, as it directly governs the ability to model the latent distribution and balance geometric plausibility with novelty.

\vspace{-0.4em}
\section{Related Work}
\vspace{-0.5em}
\paragraph{3D Visual Content Generation.}
Generative modeling of 3D structures has advanced rapidly with voxel-based autoencoders, implicit representations, and diffusion models. Early works such as voxel GANs and VAEs~\citep{wu2016learning, brock2016generative} produced coarse but plausible shapes, while point cloud and mesh models~\citep{yang2019pointflow, meshdiffusion} improved geometric fidelity. Recent diffusion-based approaches~\citep{nichol2021improved, guan20233d, dit3d} achieve strong quality and diversity for generic 3D content. However, these methods primarily target visual plausibility, whereas metamaterial geometry generation requires additional geometric regularities, such as periodic boundary consistency, symmetry, and connectivity. Thus, direct adoption of generic 3D generation is insufficient, motivating domain-specific frameworks that explicitly regulate and guide voxel geometry generation.

\paragraph{Metamaterial Geometry Generation.}
AI-driven metamaterial generation has explored multiple representations.  
Graph-based methods~\citep{UniMate, GEOLDM, uniTruss} model unit cells as nodes and edges, which is well-suited for truss-based geometries and property prediction but struggles with fine-grained geometric detail due to simplified primitives.  
2D image-based approaches~\citep{kollmann2020deep, tian2022machine, wilt2020accelerating} construct patterned planar motifs and extrude them along one axis, enabling strong in-plane performance but limited geometric variation along the extruded direction.  
Voxel-based approaches~\citep{3D-CDM, zheng2023deep, yang2024guided} offer a unified representation that can express different metamaterial geometries, such as truss, shell, and porous structures, within a single discretization. Yet, current voxel generative models often face a plausibility--novelty trade-off: staying close to the training distribution yields geometrically plausible but less novel candidates, while moving aggressively away from known samples can lead to degenerate geometries. This motivates our focus on guided voxel generation that balances geometric plausibility, novelty, and diversity.

\section{Conclusion}
\vspace{-0.5em}
In this paper, we introduced \name, a framework for voxel-based metamaterial geometry generation that combines latent space regulation with guided diffusion. RAS separates plausible geometries from perturbed degenerate ones for robust decoding, while SRR discourages generation overly close to known samples while maintaining geometric plausibility. We also introduced MetaTruss, a voxel dataset for truss-based metamaterial geometries, and a benchmark covering geometric plausibility, novelty, and diversity. Experiments on MetaTruss and MetaShell show consistent gains over voxel-based generative baselines, suggesting that \name~effectively balances these three aspects. This work supports diverse metamaterial geometry candidate generation for downstream evaluation.



\newpage
{
\small
\bibliography{ref}

@inproceedings{SLat,
  title={Structured 3d latents for scalable and versatile 3d generation},
  author={Xiang, Jianfeng and Lv, Zelong and Xu, Sicheng and Deng, Yu and Wang, Ruicheng and Zhang, Bowen and Chen, Dong and Tong, Xin and Yang, Jiaolong},
  booktitle={Proceedings of the Computer Vision and Pattern Recognition Conference},
  pages={21469--21480},
  year={2025}
}

@article{3D-CDM,
  title={Optimizing Metamaterial Inverse Design with 3D Conditional Diffusion Model and Data Augmentation},
  author={Zheng, Xiaoyang and Shiomi, Junichiro and Yamada, Takayuki},
  journal={Advanced Materials Technologies},
  pages={2500293},
  year={2025},
  publisher={Wiley Online Library}
}

@article{guan20233d,
  title={3d equivariant diffusion for target-aware molecule generation and affinity prediction},
  author={Guan, Jiaqi and Qian, Wesley Wei and Peng, Xingang and Su, Yufeng and Peng, Jian and Ma, Jianzhu},
  journal={arXiv preprint arXiv:2303.03543},
  year={2023}
}

@article{meshdiffusion,
  title={Meshdiffusion: Score-based generative 3d mesh modeling},
  author={Liu, Zhen and Feng, Yao and Black, Michael J and Nowrouzezahrai, Derek and Paull, Liam and Liu, Weiyang},
  journal={arXiv preprint arXiv:2303.08133},
  year={2023}
}

@inproceedings{xcube,
  title={Xcube: Large-scale 3d generative modeling using sparse voxel hierarchies},
  author={Ren, Xuanchi and Huang, Jiahui and Zeng, Xiaohui and Museth, Ken and Fidler, Sanja and Williams, Francis},
  booktitle={Proceedings of the IEEE/CVF conference on computer vision and pattern recognition},
  pages={4209--4219},
  year={2024}
}

@article{dit3d,
  title={Dit-3d: Exploring plain diffusion transformers for 3d shape generation},
  author={Mo, Shentong and Xie, Enze and Chu, Ruihang and Hong, Lanqing and Niessner, Matthias and Li, Zhenguo},
  journal={Advances in neural information processing systems},
  volume={36},
  pages={67960--67971},
  year={2023}
}

@article{yang2024guided,
  title={Guided diffusion for fast inverse design of density-based mechanical metamaterials},
  author={Yang, Yanyan and Wang, Lili and Zhai, Xiaoya and Chen, Kai and Wu, Wenming and Zhao, Yunkai and Liu, Ligang and Fu, Xiao-Ming},
  journal={arXiv preprint arXiv:2401.13570},
  year={2024}
}

@article{wu2016learning,
  title={Learning a probabilistic latent space of object shapes via 3d generative-adversarial modeling},
  author={Wu, Jiajun and Zhang, Chengkai and Xue, Tianfan and Freeman, Bill and Tenenbaum, Josh},
  journal={Advances in neural information processing systems},
  volume={29},
  year={2016}
}

@article{brock2016generative,
  title={Generative and discriminative voxel modeling with convolutional neural networks. arXiv 2016},
  author={Brock, A and Lim, T and Ritchie, JM and Weston, N},
  journal={arXiv preprint arXiv:1608.04236},
  volume={4232},
  year={2016}
}

@inproceedings{yang2019pointflow,
  title={Pointflow: 3d point cloud generation with continuous normalizing flows},
  author={Yang, Guandao and Huang, Xun and Hao, Zekun and Liu, Ming-Yu and Belongie, Serge and Hariharan, Bharath},
  booktitle={Proceedings of the IEEE/CVF international conference on computer vision},
  pages={4541--4550},
  year={2019}
}

@inproceedings{nichol2021improved,
  title={Improved denoising diffusion probabilistic models},
  author={Nichol, Alexander Quinn and Dhariwal, Prafulla},
  booktitle={International conference on machine learning},
  pages={8162--8171},
  year={2021},
  organization={PMLR}
}

@InProceedings{GEOLDM,
  title = 	 {Geometric Latent Diffusion Models for 3{D} Molecule Generation},
  author =       {Xu, Minkai and Powers, Alexander S and Dror, Ron O. and Ermon, Stefano and Leskovec, Jure},
  booktitle = 	 {Proceedings of the 40th ICML},
  pages = 	 {38592--38610},
  year = 	 {2023},
  volume = 	 {202},
  month = 	 {23--29 Jul},
}

@article{uniTruss,
  title={Unifying the design space and optimizing linear and nonlinear truss metamaterials by generative modeling},
  author={Zheng, Li and Karapiperis, Konstantinos and Kumar, Siddhant and Kochmann, Dennis M},
  journal={Nature Communications},
  volume={14},
  number={1},
  pages={7563},
  year={2023},
  publisher={Nature Publishing Group UK London}
}

@inproceedings{UniMate,
  title={UniMate: A Unified Model for Mechanical Metamaterial Generation, Property Prediction, and Condition Confirmation},
  author={Zhan, Wangzhi and Chen, Jianpeng and Fu, Dongqi and Zhou, Dawei},
  booktitle={ICML},
  year={2025}
}

@inproceedings{chen2025metamatbench,
  title={Metamatbench: Integrating heterogeneous data, computational tools, and visual interface for metamaterial discovery},
  author={Chen, Jianpeng and Zhan, Wangzhi and Wang, Haohui and Jia, Zian and Gan, Jingru and Zhang, Junkai and Qi, Jingyuan and Chen, Tingwei and Huang, Lifu and Chen, Muhao and others},
  booktitle={Proceedings of the 31st ACM SIGKDD Conference on Knowledge Discovery and Data Mining V. 2},
  pages={5334--5344},
  year={2025}
}

@article{zheng2023deep,
  title={Deep-learning-based inverse design of three-dimensional architected cellular materials with the target porosity and stiffness using voxelized Voronoi lattices},
  author={Zheng, Xiaoyang and Chen, Ta-Te and Jiang, Xiaoyu and Naito, Masanobu and Watanabe, Ikumu},
  journal={Science and Technology of Advanced Materials},
  volume={24},
  number={1},
  pages={2157682},
  year={2023},
  publisher={Taylor \& Francis}
}

@article{bertoldi2017flexible,
  title={Flexible mechanical metamaterials},
  author={Bertoldi, Katia and Vitelli, Vincenzo and Christensen, Johan and Van Hecke, Martin},
  journal={Nature Reviews Materials},
  volume={2},
  number={11},
  pages={1--11},
  year={2017},
  publisher={Nature Publishing Group}
}

@article{liu2011metamaterials,
  title={Metamaterials: a new frontier of science and technology},
  author={Liu, Yongmin and Zhang, Xiang},
  journal={Chemical Society Reviews},
  volume={40},
  number={5},
  pages={2494--2507},
  year={2011},
  publisher={Royal Society of Chemistry}
}

@article{MetaModulus,
  title={Exploring the property space of periodic cellular structures based on crystal networks},
  author={Lumpe, Thomas S and Stankovic, Tino},
  journal={Proceedings of the National Academy of Sciences},
  volume={118},
  number={7},
  pages={e2003504118},
  year={2021},
  publisher={National Acad Sciences}
}

@article{kollmann2020deep,
  title={Deep learning for topology optimization of 2D metamaterials},
  author={Kollmann, Hunter T and Abueidda, Diab W and Koric, Seid and Guleryuz, Erman and Sobh, Nahil A},
  journal={Materials \& Design},
  volume={196},
  pages={109098},
  year={2020},
  publisher={Elsevier}
}

@article{tian2022machine,
  title={Machine learning-based prediction and inverse design of 2D metamaterial structures with tunable deformation-dependent Poisson's ratio},
  author={Tian, Jie and Tang, Keke and Chen, Xianyan and Wang, Xianqiao},
  journal={Nanoscale},
  volume={14},
  number={35},
  pages={12677--12691},
  year={2022},
  publisher={Royal Society of Chemistry}
}

@article{mizzi2020lightweight,
  title={Lightweight mechanical metamaterials designed using hierarchical truss elements},
  author={Mizzi, Luke and Spaggiari, Andrea},
  journal={Smart Materials and Structures},
  volume={29},
  number={10},
  pages={105036},
  year={2020},
  publisher={IOP Publishing}
}

@article{song2025compressive,
  title={Compressive behavior and energy absorption of novel body-centered cubic lattice metamaterials incorporating simple cubic truss units},
  author={Song, Xuehao and Zeng, Chengjun and Hu, Junqi and Zhao, Wei and Liu, Liwu and Liu, Yanju and Leng, Jinsong},
  journal={Composite Structures},
  pages={119230},
  year={2025},
  publisher={Elsevier}
}

@article{Stiffness,
  title={Inverting the structure--property map of truss metamaterials by deep learning},
  author={Bastek, Jan-Hendrik and Kumar, Siddhant and Telgen, Bastian and Glaesener, Rapha{\"e}l N and Kochmann, Dennis M},
  journal={Proceedings of the National Academy of Sciences},
  volume={119},
  number={1},
  pages={e2111505119},
  year={2022},
  publisher={National Acad Sciences}
}

@article{nega_poisson,
  title={Hyperbolically patterned 3D graphene metamaterial with negative Poisson's ratio and superelasticity},
  author={Zhang, Qiangqiang and Xu, Xiang and Lin, Dong and Chen, Wenli and Xiong, Guoping and Yu, Yikang and Fisher, Timothy S and Li, Hui},
  journal={Advanced materials},
  volume={28},
  number={11},
  pages={2229--2237},
  year={2016}
}

@article{kadic20193d,
  title={3D metamaterials},
  author={Kadic, Muamer and Milton, Graeme W and van Hecke, Martin and Wegener, Martin},
  journal={Nature reviews physics},
  volume={1},
  number={3},
  pages={198--210},
  year={2019},
  publisher={Nature Publishing Group UK London}
}

@article{maurizi2025designing,
  title={Designing metamaterials with programmable nonlinear responses and geometric constraints in graph space},
  author={Maurizi, Marco and Xu, Derek and Wang, Yu-Tong and Yao, Desheng and Hahn, David and Oudich, Mourad and Satpati, Anish and Bauchy, Mathieu and Wang, Wei and Sun, Yizhou and others},
  journal={Nature Machine Intelligence},
  pages={1--14},
  year={2025},
  publisher={Nature Publishing Group UK London}
}

@article{wilt2020accelerating,
  title={Accelerating auxetic metamaterial design with deep learning},
  author={Wilt, Jackson K and Yang, Charles and Gu, Grace X},
  journal={Advanced Engineering Materials},
  volume={22},
  number={5},
  pages={1901266},
  year={2020},
  publisher={Wiley Online Library}
}

@article{vae,
  title={Auto-encoding variational bayes},
  author={Kingma, Diederik P and Welling, Max},
  journal={arXiv preprint arXiv:1312.6114},
  year={2013}
}

@article{ddpm,
  title={Denoising diffusion probabilistic models},
  author={Ho, Jonathan and Jain, Ajay and Abbeel, Pieter},
  journal={Advances in neural information processing systems},
  volume={33},
  pages={6840--6851},
  year={2020}
}

@inproceedings{lat_diff,
  title={High-resolution image synthesis with latent diffusion models},
  author={Rombach, Robin and Blattmann, Andreas and Lorenz, Dominik and Esser, Patrick and Ommer, Bj{\"o}rn},
  booktitle={Proceedings of the IEEE/CVF conference on computer vision and pattern recognition},
  pages={10684--10695},
  year={2022}
}

@inproceedings{contrastive,
  title={A theoretical analysis of contrastive unsupervised representation learning},
  author={Saunshi, Nikunj and Plevrakis, Orestis and Arora, Sanjeev and Khodak, Mikhail and Khandeparkar, Hrishikesh},
  booktitle={International conference on machine learning},
  pages={5628--5637},
  year={2019},
  organization={PMLR}
}

@article{coulomb,
  title={Coulomb repulsion and correlation strength in LaFeAsO from density functional anddynamical mean-field theories},
  author={Anisimov, VI and Korotin, Dm M and Korotin, MA and Kozhevnikov, AV and Kune{\v{s}}, Jan and Shorikov, AO and Skornyakov, SL and Streltsov, SV},
  journal={Journal of Physics: Condensed Matter},
  volume={21},
  number={7},
  pages={075602},
  year={2009},
  publisher={IOP Publishing}
}
\bibliographystyle{unsrt}
}


\newpage
\appendix
\onecolumn
\section{More Generation Results}
\label{app:gen_results}

\subsection{Generation Results for MetaTruss}
\begin{figure}[htbp]
    \centering
    \begin{subfigure}[b]{0.3\textwidth}
        \centering
        \includegraphics[width=\linewidth]{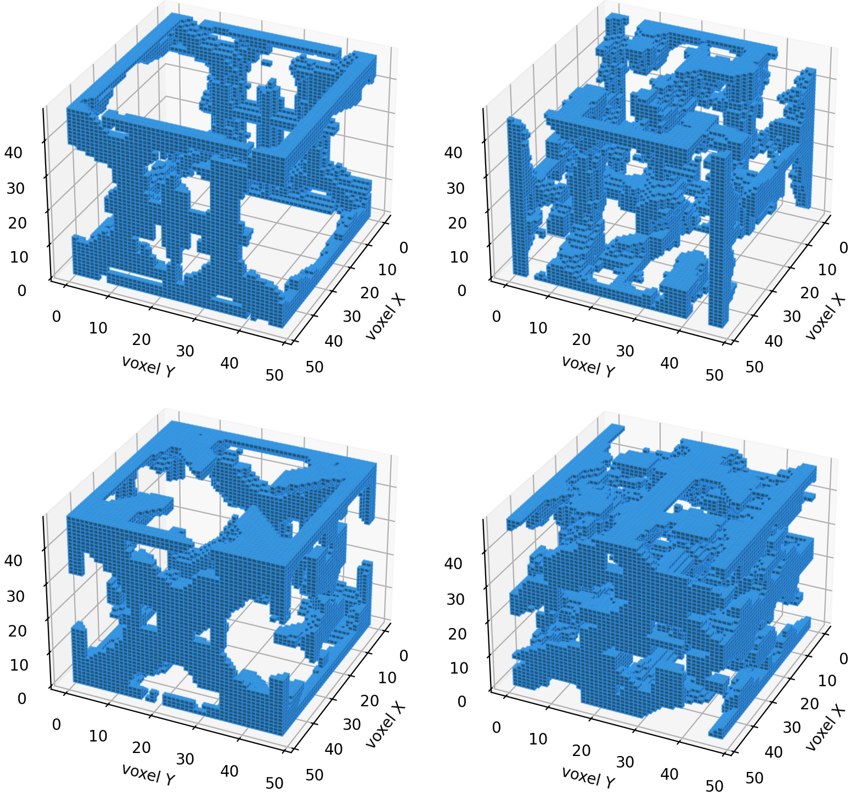}
        \caption{\name~(ours)}
        \label{fig:sub1}
    \end{subfigure}
    \hfill
    \begin{subfigure}[b]{0.3\textwidth}
        \centering
        \includegraphics[width=\linewidth]{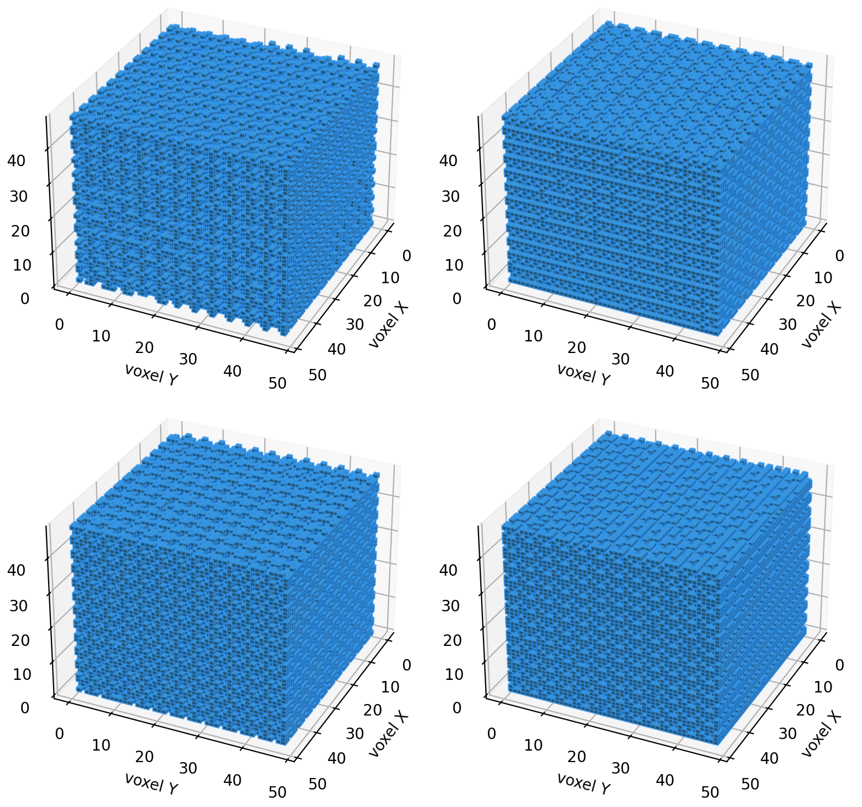}
        \caption{Dit-3D}
        \label{fig:sub2}
    \end{subfigure}
    \hfill
    \begin{subfigure}[b]{0.3\textwidth}
        \centering
        \includegraphics[width=\linewidth]{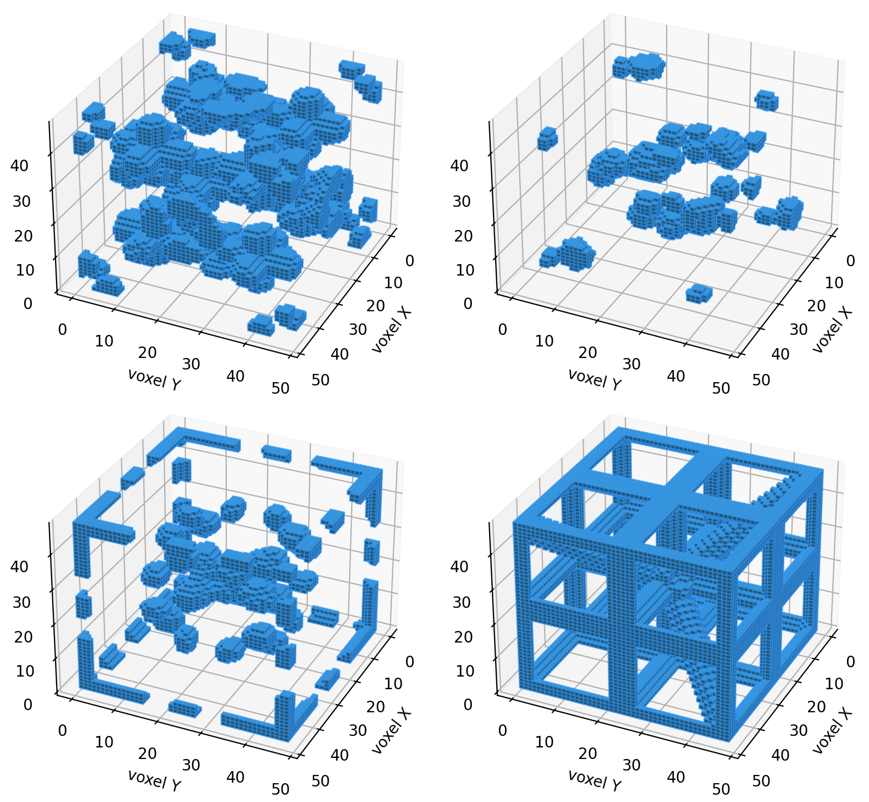}
        \caption{~\cite{yang2024guided}}
        \label{fig:sub3}
    \end{subfigure}
    
    \vspace{0.5em}
    
    \begin{subfigure}[b]{0.3\textwidth}
        \centering
        \includegraphics[width=\linewidth]{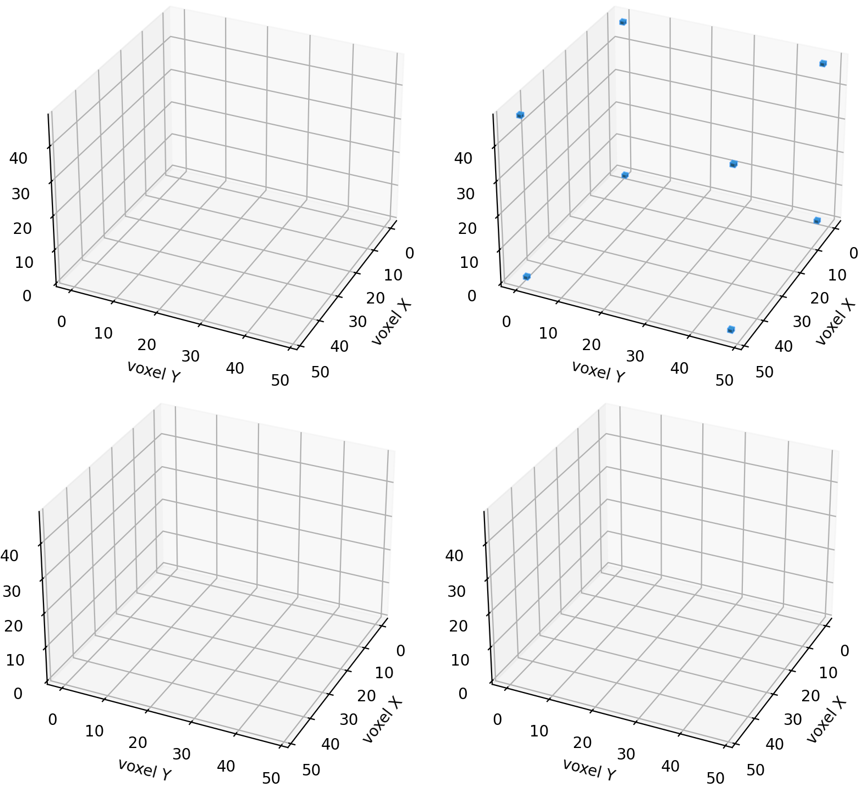}
        \caption{XCube}
        \label{fig:sub4}
    \end{subfigure}
    \hfill
    \begin{subfigure}[b]{0.3\textwidth}
        \centering
        \includegraphics[width=\linewidth]{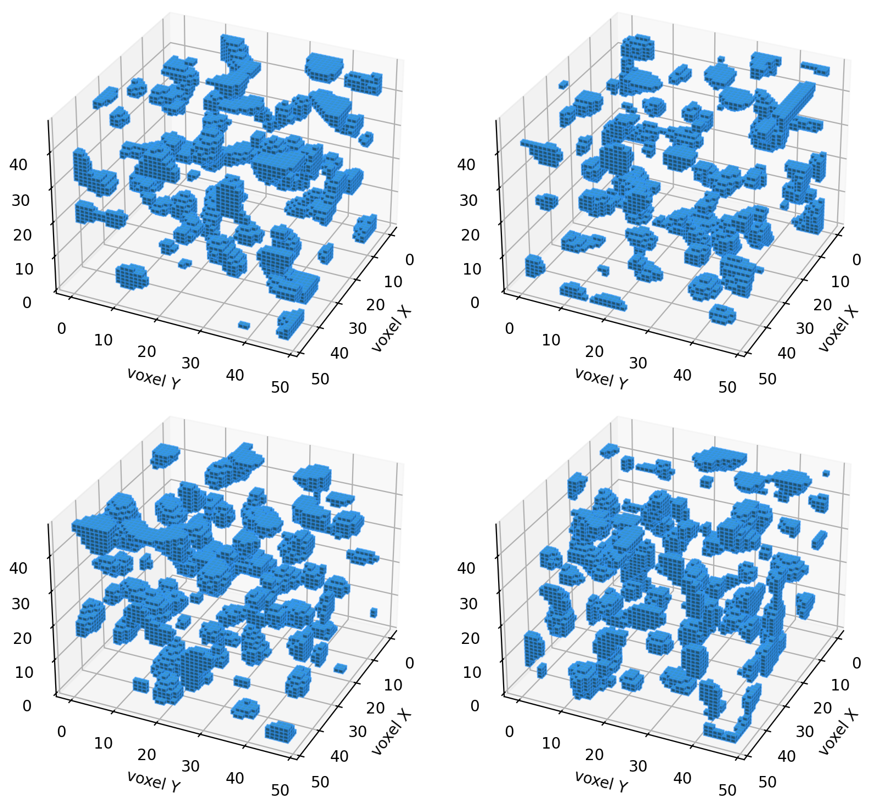}
        \caption{Trellis}
        \label{fig:sub5}
    \end{subfigure}
    \hfill
    \begin{subfigure}[b]{0.3\textwidth}
        \centering
        \includegraphics[width=\linewidth]{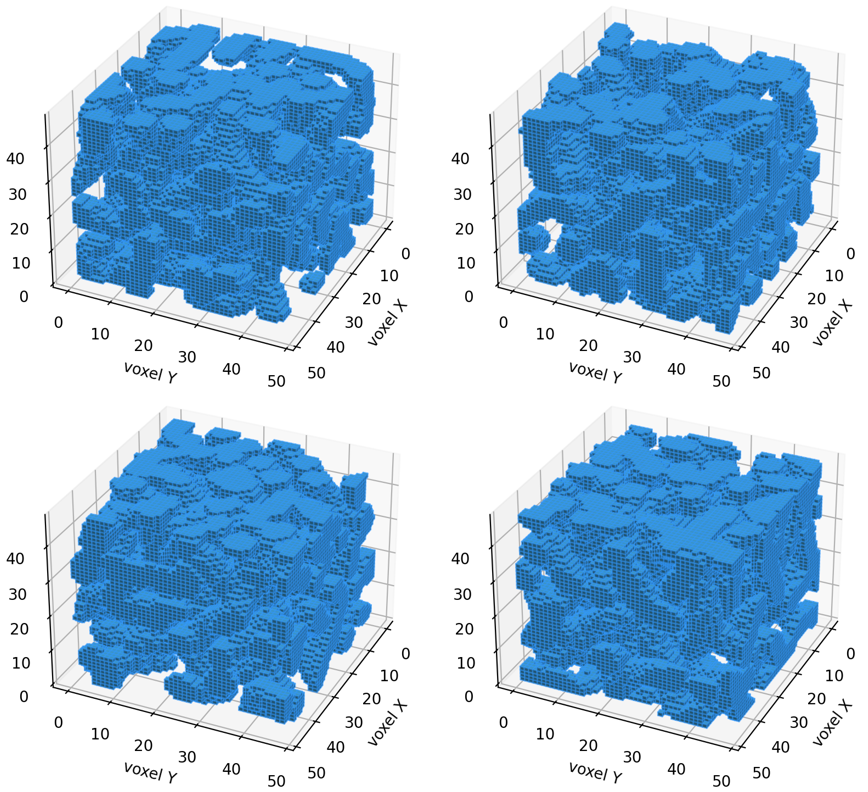}
        \caption{3D-CDM}
        \label{fig:sub6}
    \end{subfigure}
    
    \caption{Generated samples on MetaTruss with different models.}
    \label{fig:meta_truss_gen}
\end{figure}

Figure~\ref{fig:meta_truss_gen} shows some generated samples from all baselines and our model, which are trained on MetaTruss dataset.

\vspace{-1em}

\begin{figure}[htbp]
    \centering
    \begin{subfigure}[b]{0.3\textwidth}
        \centering
        \includegraphics[width=\linewidth]{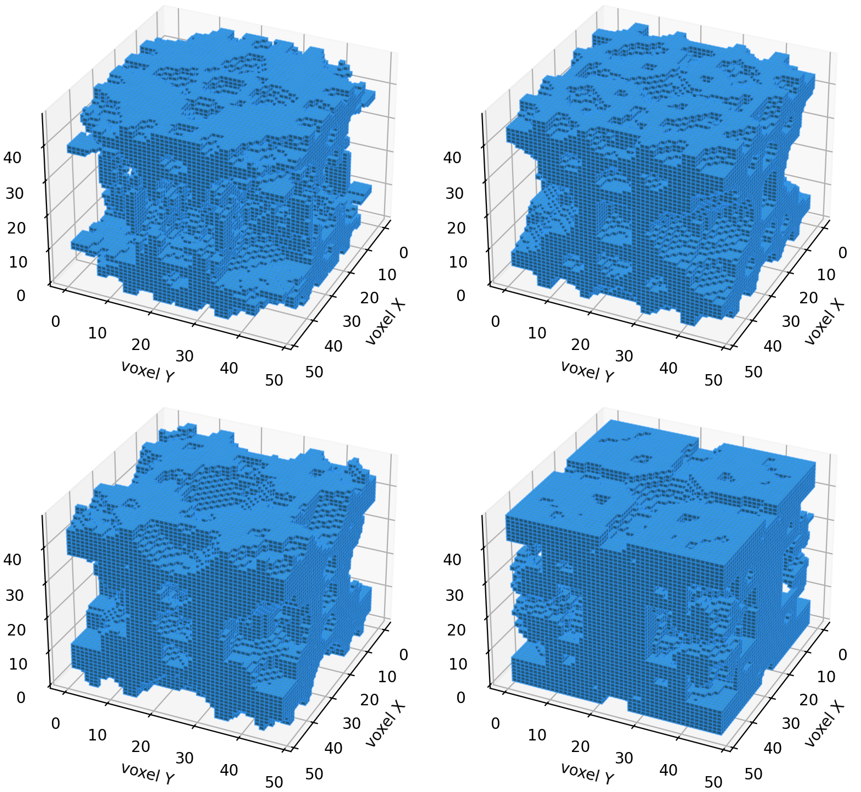}
        \caption{\name~(ours)}
        \label{fig:sub1}
    \end{subfigure}
    \hfill
    \begin{subfigure}[b]{0.3\textwidth}
        \centering
        \includegraphics[width=\linewidth]{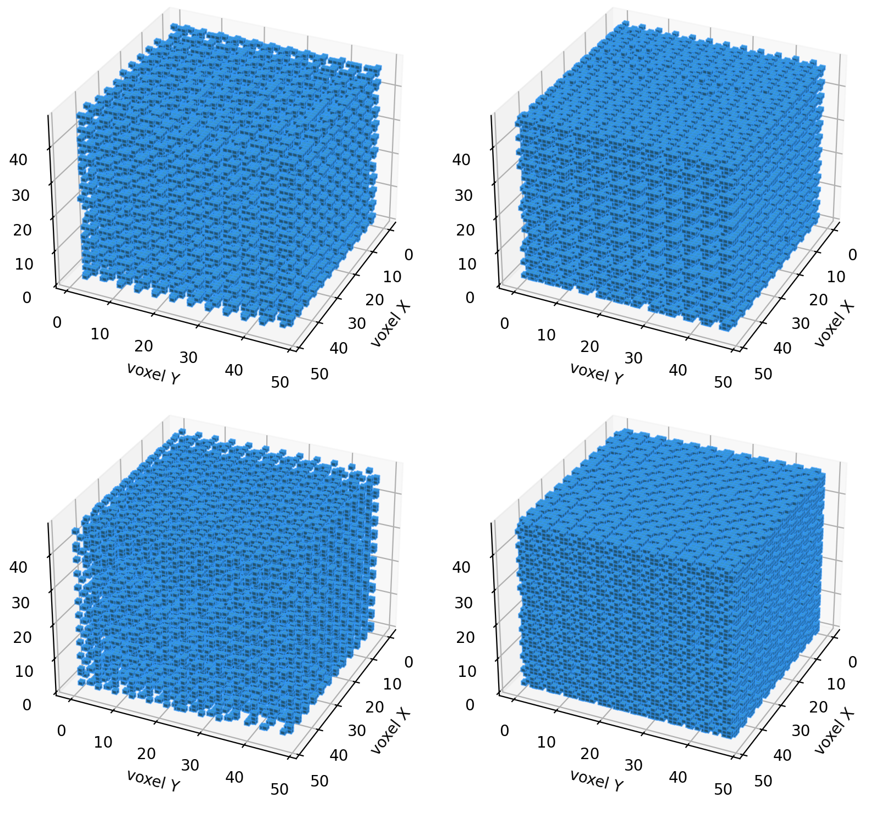}
        \caption{Dit-3D}
        \label{fig:sub2}
    \end{subfigure}
    \hfill
    \begin{subfigure}[b]{0.3\textwidth}
        \centering
        \includegraphics[width=\linewidth]{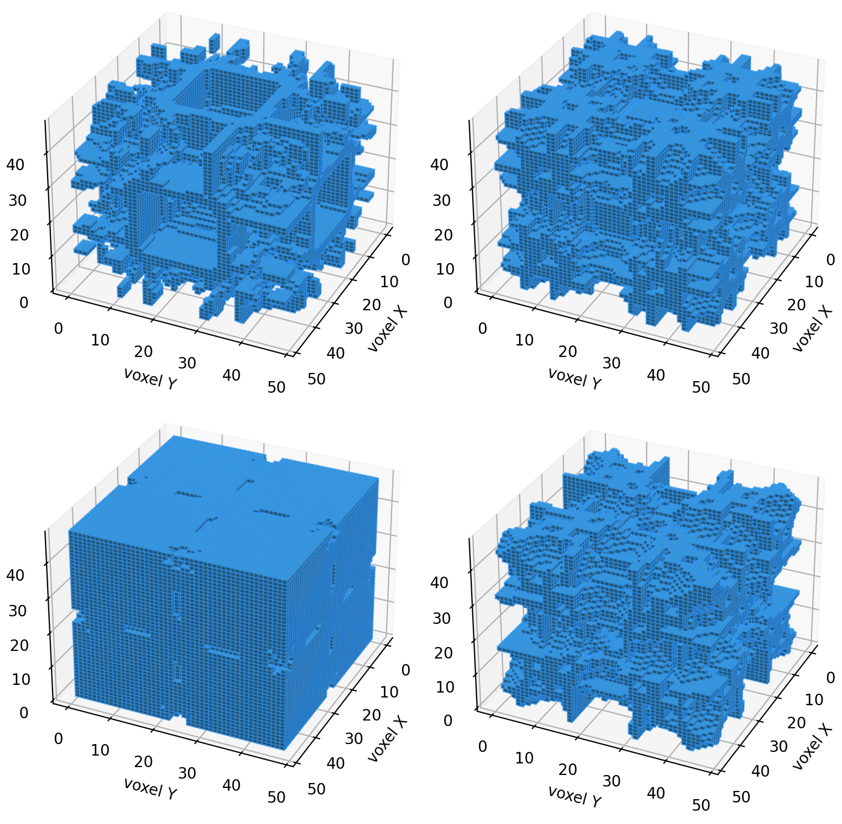}
        \caption{~\cite{yang2024guided}}
        \label{fig:sub3}
    \end{subfigure}
    
    \vspace{0.5em}
    
    \begin{subfigure}[b]{0.3\textwidth}
        \centering
        \includegraphics[width=\linewidth]{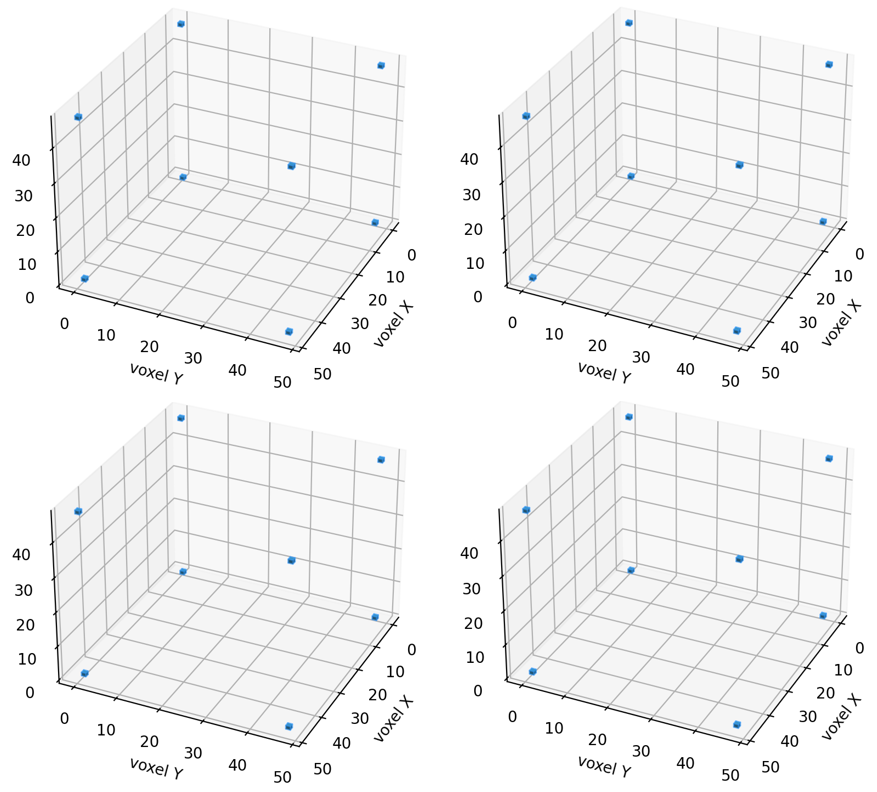}
        \caption{XCube}
        \label{fig:sub4}
    \end{subfigure}
    \hfill
    \begin{subfigure}[b]{0.3\textwidth}
        \centering
        \includegraphics[width=\linewidth]{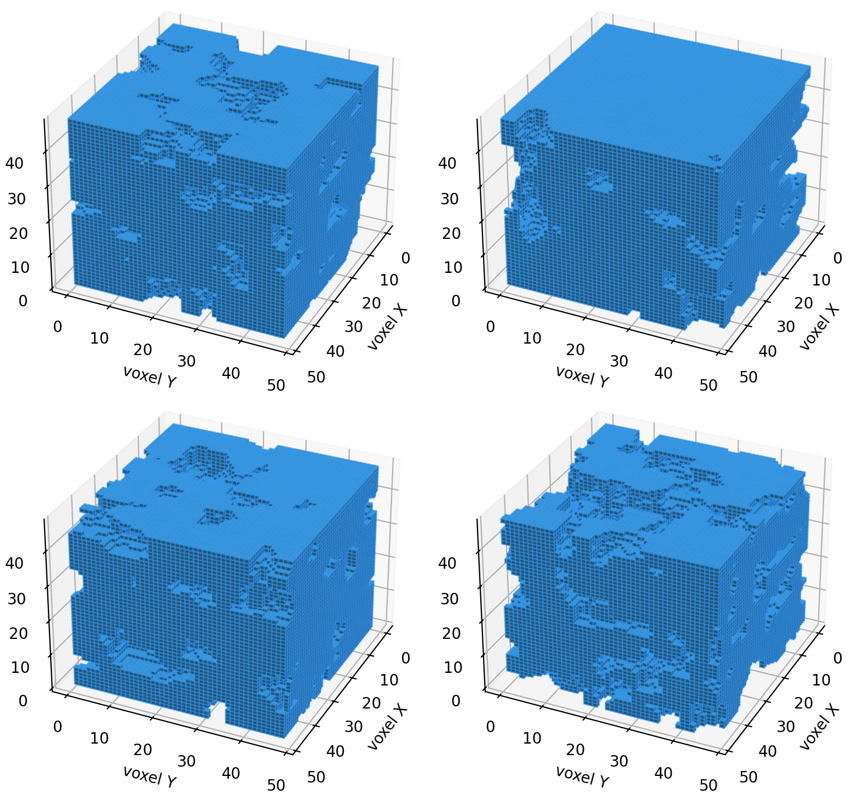}
        \caption{Trellis}
        \label{fig:sub5}
    \end{subfigure}
    \hfill
    \begin{subfigure}[b]{0.3\textwidth}
        \centering
        \includegraphics[width=\linewidth]{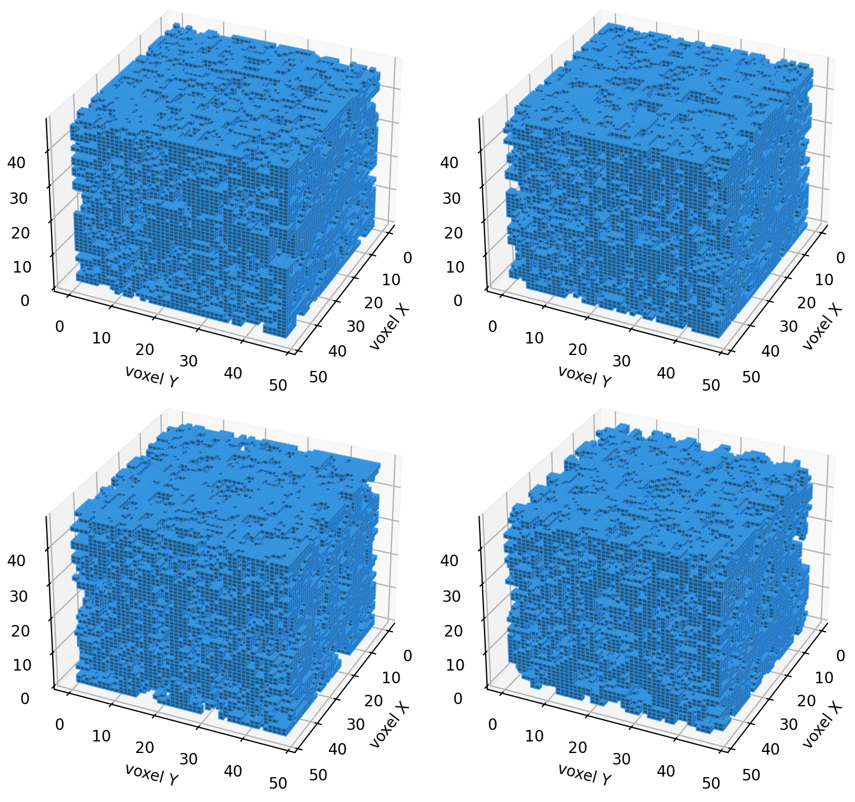}
        \caption{3D-CDM}
        \label{fig:sub6}
    \end{subfigure}
    
    \caption{Generated samples on MetaShell with different models.}
    \label{fig:meta_shell_gen}
\end{figure}

\subsection{Generation Results for MetaShell}
Figure~\ref{fig:meta_shell_gen} shows some generated samples from all baselines and our model, which are trained on MetaShell dataset.

\paragraph{Remarks.} The generation results of our~\name~have both improved novelty and genuine plausibility compared with other baselines. These results can serve as novel geometry candidates for human experts to evaluate or find inspiration from. 

\section{Details on Model Architecture and Training Scheme}
\label{app:model_detail}
\subsection{Model Architecture}
The encoder and decoder we use are two transformers of the same structure, which have 4 layers and whose model dimension is by default 128. The latent space is also set to be 128-dimensional. The voxel data are decomposed into patches with the size of $8^3$, and flattened as the input to the encoder. After the encoder $\mathcal{E}$ and before the decoder $\mathcal{D}$, there is each an 2-layer multilayer-perceptron (MLP) to resize the data to and from 128-dimensional.
\vspace{1em}\\
The latent diffusion model we use has a backbone of MLP, which has 16 layers and a model dimension of 512, with residual links connecting adjacent layers.

\subsection{Training Scheme}
The autoencoder is trained with RAS regulation. To enable this operation, we have to construct a negative dataset and combine it with the initial positive dataset. The negative data are created by noising each positive sample. We randomly select an eighth of the voxels and substitute them to void or Gaussian noise or an eighth of another positive sample, or simply add Gaussian to the initial values.
\vspace{1em}\\
The diffusion model is trained following the DDPM paradigm, and the SRR mechanism only functions in inference stage.

\section{Details on Benchmark}
\label{app:bench_detail}

\subsection{Dataset Creation and Representation Unification}
\begin{figure}
    \centering
    \includegraphics[width=0.7\linewidth]{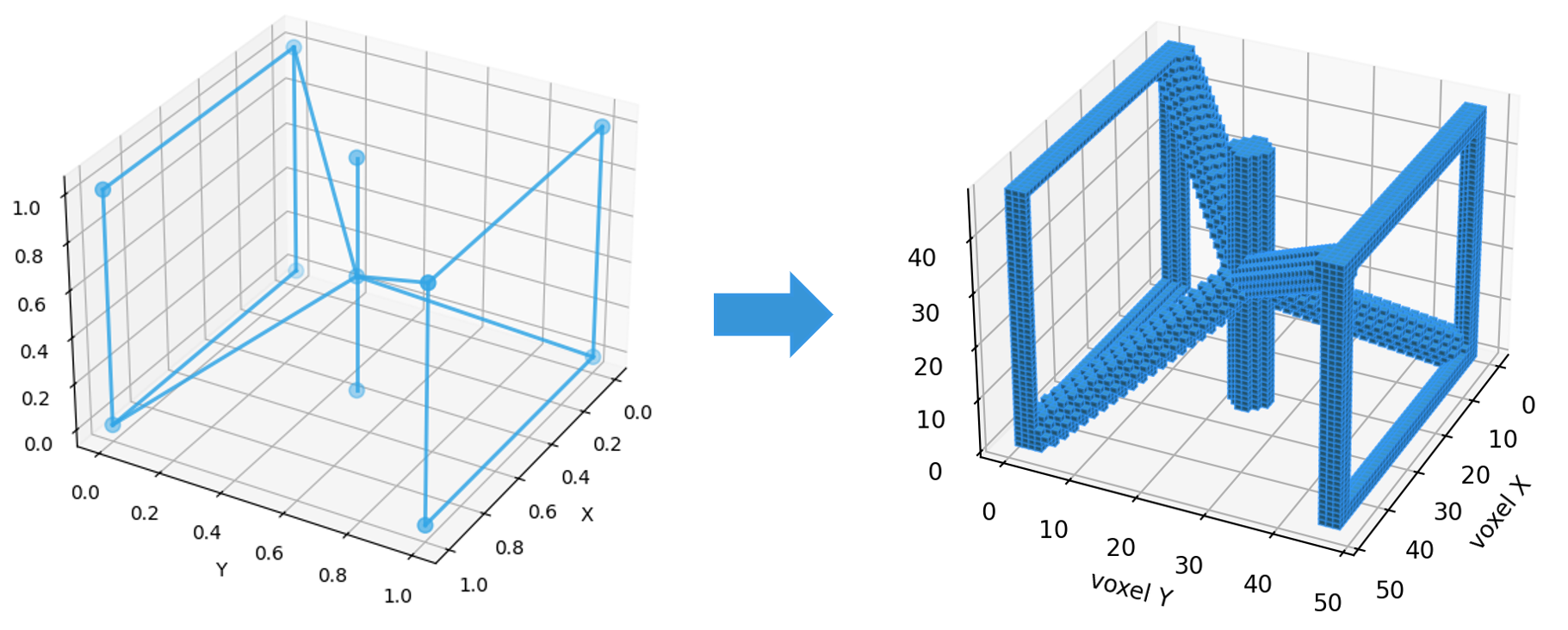}
    \caption{Data creation for MetaTruss.}
    \label{fig:data_gen}
\end{figure}

Our dataset is created based on the dataset from~\cite{MetaModulus}, which comprises over 17,000 samples in 3D graph representation (left side of Figure~\ref{fig:data_gen}). We select the first 10,000 samples from~\cite{MetaModulus} and compute whether each voxel is close enough to any 3D edge in the 3D graph. If the distance is less than a predefined radius (\eg, 0.06), then the voxel is decided to be solid, or else the voxel is void. The distance $\delta$ between a voxel's center $\boldsymbol{c}$  an edge whose endpoints are $\boldsymbol{p}_1$ and $\boldsymbol{p}_2$ is:

\begin{equation}
\delta=\frac{||(\boldsymbol{p}_1-\boldsymbol{c})\times{(\boldsymbol{p}_1-\boldsymbol{p}_2)}||}{||\boldsymbol{p}_1-\boldsymbol{p}_2||},
\end{equation}
where $\times$ means outer product. Figure~\ref{fig:data_gen} gives an example of a structure before and after the above operation.

\vspace{1em}
In the setting of our benchmark, the resolution of a voxel sample is determined to be 48. When constructing MetaTruss dataset, we directly make the shape of data to be $48^3$. The voxel data in MetaShell is initially of shape $128^3$. To unify the data representation, we use ``skimage.transform.resize" to resize the voxel data into the needed dimension with interpolation.

\subsection[Evaluation Module]{Evaluation Module}

The evaluation module of our benchmark systematically evaluates the voxel data from three aspects: geometric plausibility, novelty and diversity. For geometric plausibility, we are inspired from~\cite{chen2025metamatbench} where the symmetry, periodicity and connectivity of the generated samples are calculated. However, the benchmark in~\cite{chen2025metamatbench} is designed for graph-representation. In this paper we generalize the idea to voxel domain, and define the following three quality metrics:
\begin{equation}
S_\mathrm{sym}(\mathbf{U}^\mathrm{gen})=1-\frac{\sum_{i,j,k=1}^N|u_{i,j,k}^\mathrm{gen}-u_{N+1-i,N+1-j,N+1-k}^\mathrm{gen}|}{2|\mathbf{U}^\mathrm{gen}|},
\end{equation}

\begin{equation}
S_\mathrm{per}(\mathbf{U}^\mathrm{gen})=\frac{1}{3}\left(\frac{\mathbf{U}|_{i=1}^\mathrm{gen}\cap\mathbf{U}|_{i=N}^\mathrm{gen}}{\mathbf{U}|_{i=1}^\mathrm{gen}\cup\mathbf{U}|_{i=N}^\mathrm{gen}}+
\frac{\mathbf{U}|_{j=1}^\mathrm{gen}\cap\mathbf{U}|_{j=N}^\mathrm{gen}}{\mathbf{U}|_{j=1}^\mathrm{gen}\cup\mathbf{U}|_{j=N}^\mathrm{gen}}
+\frac{\mathbf{U}|_{k=1}^\mathrm{gen}\cap\mathbf{U}|_{k=N}^\mathrm{gen}}{\mathbf{U}|_{k=1}^\mathrm{gen}\cup\mathbf{U}|_{k=N}^\mathrm{gen}}\right),
\end{equation}

\begin{equation}
S_\mathrm{con}(\mathbf{U}^\mathrm{gen})=\frac{\max\limits_i{|\mathbf{C}_i|}}{|\mathbf{U}^\mathrm{gen}|},
\end{equation}

where $S_\mathrm{sym}$ measure the central symmetry degree, $S_\mathrm{per}$ measures the periodicity degree, and $S_\mathrm{con}$ measures the connectivity degree; $\mathbf{U}^\mathrm{gen}$ is a generated sample in voxel representation, $\mathbf{C}_i$ is the $i$th cluster of connected voxels in $\mathbf{U}^\mathrm{gen}$, $u_{i,j,k}^\mathrm{gen}$ is a voxel in $\mathbf{U}^\mathrm{gen}$ whose indices are $i,j,k$, $\mathbf{U}|_{i=1}^\mathrm{gen}$ is a slice of voxels in $\mathbf{U}^\mathrm{gen}$ whose the index along x axis is $i=1$, and $|\mathbf{U}^\mathrm{gen}|$ is the number of solid voxels in $\mathbf{U}^\mathrm{gen}$.
\vspace{1em}\\
For novelty we propose a distribution-based novelty score:
\begin{equation}
S_\mathrm{nov}(\mathbf{U}^\mathrm{gen};\mathcal{U_\mathrm{train}})=1-\frac{\mathbf{U}^\mathrm{gen}\cap\mathbf{U}_\mathrm{NN}^\mathrm{train}}{\mathbf{U}^\mathrm{gen}\cup\mathbf{U}_\mathrm{NN}^\mathrm{train}},
\end{equation}
where $\mathbf{U}_\mathrm{NN}^\mathrm{train}$ is the nearest neighbor of $\mathbf{U}^\mathrm{gen}$ in the training dataset $\mathcal{U}^\mathrm{train}$.
\vspace{1em}\\
For diversity we propose a distribution-based diversity score:
\begin{equation}
S_\mathrm{div}(\mathcal{U}^\mathrm{gen};\mathcal{U}^\mathrm{train})=\frac{|\mathcal{L}|}{|\mathcal{U}^\mathrm{gen}|},
\end{equation}

\begin{equation}
\mathcal{L}=\left\{l_i|l_i=\arg\max\limits_{l'\in\{1,2,\cdots,|\mathcal{U}^\mathrm{train}|\}}\frac{\mathbf{U}_i^\mathrm{gen}\cap\mathbf{U}_{l'}^\mathrm{train}}{\mathbf{U}_i^\mathrm{gen}\cup\mathbf{U}_{l'}^\mathrm{train}}, i\in\{1,2,\cdots,|\mathcal{U}^\mathrm{gen}|\}\right\},
\end{equation}

where $\mathbf{U}_i^\mathrm{gen}$ is the $i$th elements in the set of generated samples $\mathcal{U}^\mathrm{gen}$, $\mathbf{U}_{l'}^\mathrm{train}$ is the $l'$th elements in $\mathcal{U}^\mathrm{train}$.

\section{More Analytical Experiments}
\subsection{Ablation Results on MetaShell Dataset}
\label{app:shell_ablation}
In this section we provide some extra ablation results conducted on the MetaShell dataset.

\begin{table}[h]
\centering
\caption{Ablation on RAS regulation and SRR diffusion.}
\label{tab:SRR_effect_shell}
\setlength{\tabcolsep}{4pt}
\small
\begin{tabular}{@{}lcccccc@{}}
\toprule
\multirow{2}{*}{Approaches} & \multicolumn{4}{c}{Geometric Plausibility Scores} & Novelty Score & Diversity Score\\ 
\cmidrule(lr){2-5} \cmidrule(lr){6-6} \cmidrule(lr){7-7}
& $S_\mathrm{sym}$ $\uparrow$ & $S_\mathrm{per}$ $\uparrow$ & $S_\mathrm{con}$ $\uparrow$ & Mean $\uparrow$ & $S_\mathrm{nov}$ $\uparrow$ & $S_\mathrm{div}$ $\uparrow$ \\ 
\midrule
Case 1 (RAS + vanilla DDPM)   & \textbf{0.935} & \textbf{0.884} & \underline{0.930} & \underline{0.916} & \underline{0.305} & \underline{0.625} \\
Case 2 (w/o reg + SRR Diff.) & 0.910 & \textbf{0.795} & 0.842 & 0.849 & 0.237 & 0.477 \\
Case 3 (full framework)     & \underline{0.923} & \underline{0.856} & \textbf{0.978} & \textbf{0.919} & \textbf{0.380} & \textbf{0.783} \\
\bottomrule
\end{tabular}
\end{table}
\vspace{-1em}

\begin{table}[h]
\centering
\caption{Ablation on model capacity. Param. Num. denotes parameter number.}
\label{tab:sense_shell}
\setlength{\tabcolsep}{4pt}
\small
\begin{tabular}{@{}lcccccc@{}}
\toprule
\multirow{2}{*}{Approaches} & \multicolumn{4}{c}{Geometric Plausibility Scores} & Novelty Score & Diversity Score\\ 
\cmidrule(lr){2-5} \cmidrule(lr){6-6} \cmidrule(lr){7-7}
& $S_\mathrm{sym}$ $\uparrow$ & $S_\mathrm{per}$ $\uparrow$ & $S_\mathrm{con}$ $\uparrow$ & Mean $\uparrow$ & $S_\mathrm{nov}$ $\uparrow$ & $S_\mathrm{div}$ $\uparrow$ \\ 
\midrule
Increase AE Param. Num.   & \underline{0.915} & \textbf{0.858} & \textbf{0.985} & \textbf{0.919} & 0.362 & 0.711 \\
Decrease AE Param. Num.   & 0.894 & 0.823 & 0.963 & 0.893 & 0.363 & 0.742 \\
Increase diff. Param. Num.& 0.920 & 0.810 & 0.953 & 0.894 & \textbf{0.389} & \underline{0.781} \\
Decrease diff. Param. Num.& 0.907 & 0.852 & 0.956 & \underline{0.905} & 0.359 & 0.775 \\
Original setting          & \textbf{0.923} & \underline{0.856} & \underline{0.978} & \textbf{0.919} & \underline{0.380} & \textbf{0.783} \\
\bottomrule
\end{tabular}
\end{table}

\subsection{Time Efficiency Comparison}
In this section we compare the efficiency of metamaterial generation of different models. All models are set to generate 1,000 metamaterial structures, and Table~\ref{tab:time_eff} shows the time consumed for this task.
\begin{table}[h]
\centering
\caption{Time efficiency comparison.}
\label{tab:time_eff}
\begin{tabular}{@{}ll@{}}
\toprule
Approaches & Time (s) \\
\midrule
DiT-3D (\cite{dit3d})            & 326 \\
Y. Yang et al. (\cite{yang2024guided}) & 28 \\
XCube (\cite{xcube})             & 94 \\
Trellis (\cite{SLat})            & 5  \\
3D-CDM (\cite{3D-CDM})           & 813  \\
\name~(ours)                     & 31 \\
\bottomrule
\end{tabular}
\end{table}

\section{Limitations and Broader Impact}
\label{app:limitations_impact}

\paragraph{Limitations.}
This work focuses on two representative voxel-based metamaterial families, truss-based and shell-based structures. Broader design families, physics-based validation, and higher-resolution generation remain valuable extensions to further strengthen the benchmark and framework.

\paragraph{Broader Impact.}
This work may accelerate metamaterial discovery by reducing manual trial-and-error and supporting applications such as lightweight structures, energy absorption, vibration isolation, soft robotics, and biomedical scaffolds. Generated structures should be treated as design candidates and validated by domain experts before safety-critical or real-world deployment.


\newpage
\section*{NeurIPS Paper Checklist}

\begin{enumerate}

\item {\bf Claims}
    \item[] Question: Do the main claims made in the abstract and introduction accurately reflect the paper's contributions and scope?
    \item[] Answer: \answerYes{} 
    \item[] Justification: The abstract and introduction clearly state the proposed framework, benchmark contribution, evaluation metrics, and experimental scope. The claims are supported by the main experimental results in Section~\ref{tab:effectiveness} and the ablation studies.

\item {\bf Limitations}
    \item[] Question: Does the paper discuss the limitations of the work performed by the authors?
    \item[] Answer: \answerYes{} 
    \item[] Justification: Limitations are discussed in Appendix~\ref{app:limitations_impact}.

\item {\bf Theory assumptions and proofs}
    \item[] Question: For each theoretical result, does the paper provide the full set of assumptions and a complete (and correct) proof?
    \item[] Answer: \answerNA{} 
    \item[] Justification: The paper does not present formal theoretical results or theorem-style proofs.

    \item {\bf Experimental result reproducibility}
    \item[] Question: Does the paper fully disclose all the information needed to reproduce the main experimental results of the paper to the extent that it affects the main claims and/or conclusions of the paper (regardless of whether the code and data are provided or not)?
    \item[] Answer: \answerYes{} 
    \item[] Justification: The paper describes the benchmark construction, evaluation metrics, model components, baselines, and experimental protocol. Code is released through an anonymized repository.

\item {\bf Open access to data and code}
    \item[] Question: Does the paper provide open access to the data and code, with sufficient instructions to faithfully reproduce the main experimental results, as described in supplemental material?
    \item[] Answer: \answerYes{} 
    \item[] Justification: The paper provides an anonymized code repository and describes the construction of the MetaTruss dataset and the use of MetaShell.

\item {\bf Experimental setting/details}
    \item[] Question: Does the paper specify all the training and test details (e.g., data splits, hyperparameters, how they were chosen, type of optimizer) necessary to understand the results?
    \item[] Answer: \answerYes{} 
    \item[] Justification: The paper specifies the datasets, baselines, evaluation metrics, model architecture, and training setup.

\item {\bf Experiment statistical significance}
    \item[] Question: Does the paper report error bars suitably and correctly defined or other appropriate information about the statistical significance of the experiments?
    \item[] Answer: \answerNo{} 
    \item[] Justification: The paper reports results across two datasets and multiple baselines though does not currently include error bars. This is mainly due to the high computational cost of repeatedly training 3D generative models and baselines.

\item {\bf Experiments compute resources}
    \item[] Question: For each experiment, does the paper provide sufficient information on the computer resources (type of compute workers, memory, time of execution) needed to reproduce the experiments?
    \item[] Answer: \answerYes{} 
    \item[] Justification: The paper reports the main compute resource in the ``Overall Comparison'' subsection.
    
\item {\bf Code of ethics}
    \item[] Question: Does the research conducted in the paper conform, in every respect, with the NeurIPS Code of Ethics \url{https://neurips.cc/public/EthicsGuidelines}?
    \item[] Answer: \answerYes{} 
    \item[] Justification: The work uses scientific structure datasets and computational experiments, does not involve private personal data or human subjects, and is released anonymously for double-blind review. We have reviewed the NeurIPS Code of Ethics and believe the work conforms to it.

\item {\bf Broader impacts}
    \item[] Question: Does the paper discuss both potential positive societal impacts and negative societal impacts of the work performed?
    \item[] Answer: \answerYes{} 
    \item[] Justification: Broader impact is discussed in Appendix~\ref{app:limitations_impact}
    
\item {\bf Safeguards}
    \item[] Question: Does the paper describe safeguards that have been put in place for responsible release of data or models that have a high risk for misuse (e.g., pre-trained language models, image generators, or scraped datasets)?
    \item[] Answer: \answerNA{} 
    \item[] Justification: The paper does not release high-risk models. The released assets are intended for scientific metamaterial generation and evaluation.

\item {\bf Licenses for existing assets}
    \item[] Question: Are the creators or original owners of assets (e.g., code, data, models), used in the paper, properly credited and are the license and terms of use explicitly mentioned and properly respected?
    \item[] Answer: \answerYes{} 
    \item[] Justification: The paper cites the original sources of existing datasets, models, and baselines used in the experiments. We respect the licenses and terms of use of all existing assets used in this work.

\item {\bf New assets}
    \item[] Question: Are new assets introduced in the paper well documented and is the documentation provided alongside the assets?
    \item[] Answer: \answerYes{} 
    \item[] Justification: The paper introduces the MetaTruss dataset and provides details on its construction. Documentation and code are provided through the anonymized repository.

\item {\bf Crowdsourcing and research with human subjects}
    \item[] Question: For crowdsourcing experiments and research with human subjects, does the paper include the full text of instructions given to participants and screenshots, if applicable, as well as details about compensation (if any)? 
    \item[] Answer: \answerNA{} 
    \item[] Justification: The paper does not involve crowdsourcing, user studies, or research with human subjects.

\item {\bf Institutional review board (IRB) approvals or equivalent for research with human subjects}
    \item[] Question: Does the paper describe potential risks incurred by study participants, whether such risks were disclosed to the subjects, and whether Institutional Review Board (IRB) approvals (or an equivalent approval/review based on the requirements of your country or institution) were obtained?
    \item[] Answer: \answerNA{} 
    \item[] Justification: The paper does not involve crowdsourcing, user studies, or research with human subjects, so IRB approval is not applicable.

\item {\bf Declaration of LLM usage}
    \item[] Question: Does the paper describe the usage of LLMs if it is an important, original, or non-standard component of the core methods in this research? Note that if the LLM is used only for writing, editing, or formatting purposes and does \emph{not} impact the core methodology, scientific rigor, or originality of the research, declaration is not required.
    \item[] Answer: \answerNA{} 
    \item[] Justification: The core method development in this research does not involve LLMs as important, original, or non-standard components.
    \item[] Guidelines:

\end{enumerate}

\end{document}